\documentclass[sigconf]{acmart}
\renewcommand\footnotetextcopyrightpermission[1]{}

\usepackage{enumitem}
\usepackage{multirow}
\usepackage{graphicx}
\usepackage{makecell}
\AtBeginDocument{%
  \providecommand\BibTeX{{%
    Bib\TeX}}}

\setcopyright{acmlicensed}
\copyrightyear{2025}
\acmYear{2025}
\acmDOI{XXXXXXX.XXXXXXX}
\acmConference[KDD '25]{ACM SIGKDD Conference on Knowledge Discovery and Data Mining}{August 03--07,
  2025}{Toronto, ON, Canada}
\acmISBN{978-1-4503-XXXX-X/2018/06}

\begin{document}

\title{UrbanPlanBench: A Comprehensive Urban Planning Benchmark for Evaluating Large Language Models}

\author{Yu Zheng}
\email{y_zheng19@mails.tsinghua.edu.cn}
\affiliation{%
  \institution{Tsinghua University}
  \city{Beijing}
  \country{China}
}

\author{Longyi Liu}
\email{liulongyi22@mails.ucas.ac.cn}
\affiliation{%
  \institution{University of Chinese Academy of Sciences}
  \city{Beijing}
  \country{China}
}

\author{Yuming Lin}
\email{linyuming9@foxmail.com}
\affiliation{%
  \institution{Tsinghua University}
  \city{Beijing}
  \country{China}
}

\author{Jie Feng}
\email{fengjie@tsinghua.edu.cn}
\affiliation{%
  \institution{Tsinghua University}
  \city{Beijing}
  \country{China}
}

\author{Guozhen Zhang}
\email{zhangguozhen@tsingroc.com}
\affiliation{%
  \institution{TsingRoc.ai}
  \city{Beijing}
  \country{China}
}

\author{Depeng Jin}
\email{jindp@tsinghua.edu.cn}
\affiliation{%
  \institution{Tsinghua University}
  \city{Beijing}
  \country{China}
}

\author{Yong Li}
\email{liyong07@tsinghua.edu.cn}
\affiliation{%
  \institution{Tsinghua University}
  \city{Beijing}
  \country{China}
}

\renewcommand{\shortauthors}{Zheng et al.}

\begin{abstract}
  The advent of Large Language Models (LLMs) holds promise for revolutionizing various fields traditionally dominated by human expertise.
  Urban planning, a professional discipline that fundamentally shapes our daily surroundings, is one such field heavily relying on multifaceted domain knowledge and experience of human experts.
  The extent to which LLMs can assist human practitioners in urban planning remains largely unexplored.
  In this paper, we introduce a comprehensive benchmark, UrbanPlanBench, tailored to evaluate the efficacy of LLMs in urban planning, which encompasses fundamental principles, professional knowledge, and management and regulations, aligning closely with the qualifications expected of human planners. 
  Through extensive evaluation, we reveal a significant imbalance in the acquisition of planning knowledge among LLMs, with even the most proficient models falling short of meeting professional standards.
  For instance, we observe that 70\% of LLMs achieve subpar performance in understanding planning regulations compared to other aspects.
  Besides the benchmark, we present the largest-ever supervised fine-tuning (SFT) dataset, UrbanPlanText, comprising over 30,000 instruction pairs sourced from urban planning exams and textbooks.
  Our findings demonstrate that fine-tuned models exhibit enhanced performance in memorization tests and comprehension of urban planning knowledge, while there exists significant room for improvement, particularly in tasks requiring domain-specific terminology and reasoning.
  By making our benchmark, dataset, and associated evaluation and fine-tuning toolsets publicly available at \url{https://github.com/tsinghua-fib-lab/PlanBench}, we aim to catalyze the integration of LLMs into practical urban planning, fostering a symbiotic collaboration between human expertise and machine intelligence.
\end{abstract}

\maketitle

\section{Introduction}

Recent breakthroughs in Large Language Models (LLMs)~\citep{touvron2023llama,zeng2023glm-130b} have showcased remarkable capabilities in generating text, reasoning, and knowledge QA, unlocking a plethora of applications ranging from chatbots~\citep{chatgpt} to programming copilots~\citep{chen2021evaluating}. 
Besides general-purpose evaluation, assessing their capabilities in specialized domains is crucial for understanding the real-world impact of LLMs~\citep{chen2023bioinfo,koncel2023bizbench,fei2023lawbench,liu2024mathbench}.
In this paper, we focus on one critical field, urban planning, which stands as a cornerstone in shaping modern city life, yielding profound influence on over 4 billion urban residents worldwide. 
Urban planning is a complex endeavor that intertwines various disciplines, demanding a deep understanding of domain knowledge. 
Despite the advent of technological advancements, the field continues to heavily rely on the expertise and experience of human planners. 
For instance, human planners devote substantial time to tasks such as planning text management, review, and assessment~\citep{zhu2024plangpt}. 
Moreover, the limitations inherent in human experience often lead to errors and inefficiencies in planning outcomes~\citep{zheng2023spatial}.

Notably, the integration of LLMs in urban planning contexts has emerged as a promising avenue, leveraging their pre-trained world knowledge to tackle complex computational tasks~\citep{zhou2024large,fu2024towards,xu2023urban,zhu2024plangpt,li2024urbangpt}. 
However, the inherent challenges of hallucination and vagueness present significant hurdles, particularly when addressing specialized problems within urban planning~\citep{zhang2023siren}. 
While various benchmarks such as SuperGLUE~\cite{wang2019superglue}, BIG-BENCH~\citep{srivastava2022beyond} and C-Eval~\citep{huang2023ceval} have been proposed to evaluate LLM effectiveness in understanding and solving intricate tasks, the absence of dedicated benchmarks for urban planning restricts our ability to quantify the extent to which LLMs acquire specialized knowledge and their potential to enhance the productivity of human planners. 

As a human-centered field, matching the performance of human planners marks a milestone for LLMs and signifies their mastery of urban planning capabilities.
In alignment with the rigorous standards set by the certified urban planner qualification examination in China, we introduce UrbanPlanBench, a comprehensive benchmark designed to evaluate LLMs across various perspectives of urban planning, including fundamental principles, professional knowledge, and management and regulations. 
The benchmark mirrors the latest available examination standards as of 2022, enabling a comparative analysis between LLM and human planners, shedding lights on whether current general-purpose LLMs attain a human-level understanding of urban planning. 
Leveraging this benchmark, we scrutinize recent open-source LLMs, including LLaMA1/2/3/3.1~\citep{touvron2023llama,llama}, Gemma1/2~\cite{team2024gemma,team2024gemma2}, ChatGLM3/4~\cite{glm2024chatglm}, Baichuan2~\citep{baichuan2023baichuan2}, Qwen1.5/2~\citep{qwen,yang2024qwen2}, and Yi~\citep{young2024yi}, as well as commercial LLMs like ChatGPT 3.5/4o, to assess their acquisition of planning skills.
Additionally, we also evaluate the effect of prompting techniques for LLMs including chain of thought (COT)~\cite{wei2022chain} and retrieval augmented generation (RAG)~\cite{lewis2020retrieval,gao2023retrieval}.

Supervised fine-tuning (SFT) stands as a prevalent method to build domain-specific LLMs.
However, to our knowledge, there are currently no off-the-shelf resources for fine-tuning LLMs specifically for urban planning. 
This gap arises from the significant disparity between the distribution of urban planning knowledge in descriptive texts and the required form of SFT data, which necessitates sample pairs comprising instructions and responses.
To bridge this gap, we further introduce UrbanPlanText, the largest-ever dataset tailored for SFT of LLMs in urban planning. 
Comprising over 30,000 instruction pairs derived from textbooks and past exams, UrbanPlanText serves as a comprehensive collection of specialized urban planning contents.

We conduct extensive experiments to assess current advanced LLMs on UrbanPlanBench.
While LLMs demonstrate significantly better performance than random guessing, there remains large room for improvement, indicating a limited mastery of urban planning skills.
Notably, most of the current LLMs can not surpass the certification bar of the urban planner qualification examination, which roughly represents the top 10\% proficiency level of human planners.
Additionally, our analysis highlights an imbalance in LLM performance across three key urban planning perspectives: they exhibit greater proficiency in understanding planning principles and knowledge but tend to falter in memorizing regulations, leading to factual errors.
Moreover, we find that finetuning LLMs with UrbanPlanText can effectively enhance their ability to answer urban planning-related questions.
By introducing both UrbanPlanBench and UrbanPlanText, we aim to facilitate the seamless integration of LLMs into practical urban planning workflows, thereby lowering entry barriers for practitioners and enabling them to fully leverage advanced AI tools in their work.

\begin{figure}[t]
  \centering
  \includegraphics[width=\linewidth]{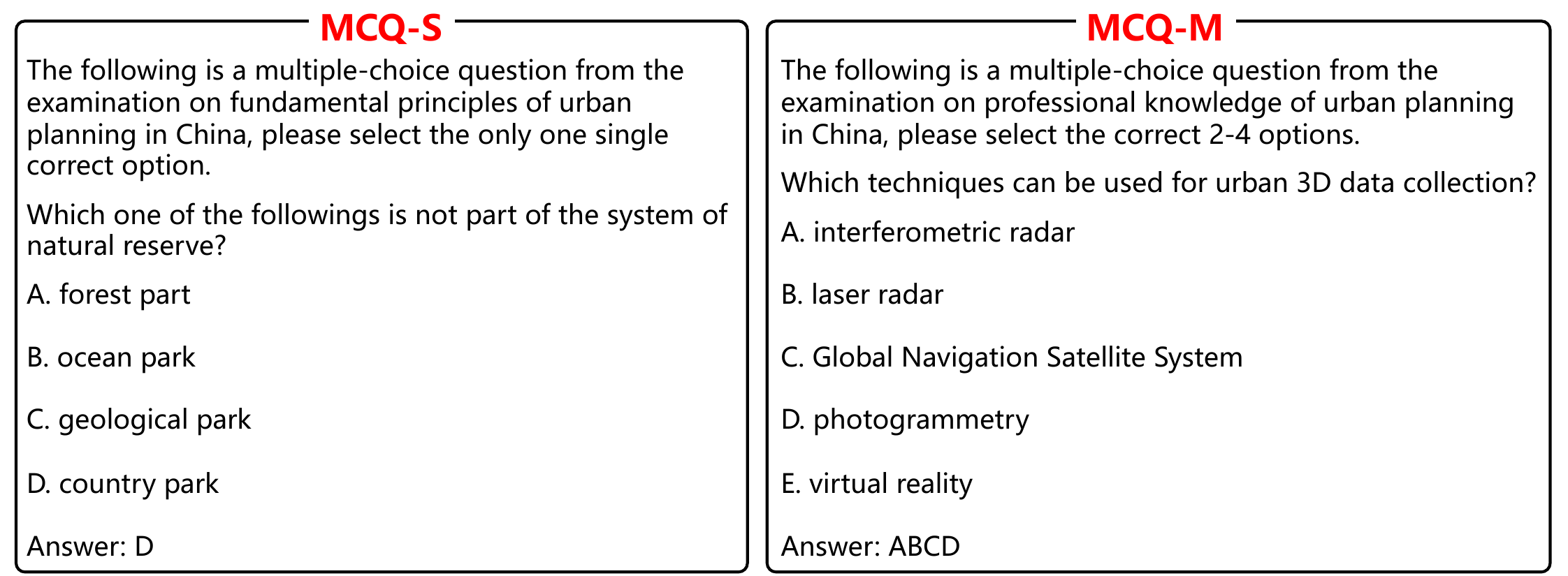}
    \vspace{-10px}
  \caption{Example questions of UrbanPlanBench.
  MCQ-S has four options where only one option is correct.
  MCQ-M is more challenging, featuring two to four correct options from a total of five options. Contents are translated from Chinese to English.}
  \label{fig::benchmark}
    \vspace{-20px}
\end{figure}

\section{Benchmarking LLMs on Urban Planning}

\begin{table*}[h]
\centering
\caption{Accuracy (\%) of LLMs on three subjects of UrbanPlanBench. S and M indicate MCQs with one single correct answer and multiple correct answers, respectively. Full represents the overall accuracy of both types of MCQs.}
\label{tab::benchmark}
\begin{tabular}{c|ccc|ccc|ccc}
\toprule
\multirow{2}{*}{\textbf{Model}} & \multicolumn{3}{c|}{\textbf{S1}} & \multicolumn{3}{c|}{\textbf{S2}} & \multicolumn{3}{c}{\textbf{S3}} \\
                                & \textbf{S}   & \textbf{M}  & \textbf{Full}  & \textbf{S}   & \textbf{M}  & \textbf{Full}  & \textbf{S}    & \textbf{M}    & \textbf{Full}   \\
\midrule
Random                          & 25.0      & 4.0       & 20.8      & 25.0      & 4.0       & 20.8      & 25.0        & 4.0        & 20.8       \\
\hline
LLaMA-7B        & 28.8 & 15.0 & 26.0 & 27.5 & 5.0  & 23.0 & 21.3 & 0.0  & 17.0  \\ 
LLaMA-13B       & 25.0 & 15.0 & 23.0 & 26.3 & 5.0  & 23.0 & 23.8 & 0.0  & 19.0  \\ 
LLaMA-30B       & 26.3 & 15.0 & 24.0 & 25.0 & 5.0  & 22.0 & 32.5 & 0.0  & 26.0  \\ 
LLaMA2-13B      & 27.5 & 15.0 & 25.0 & 28.8 & 5.0  & 24.0 & 20.0 & 0.0  & 16.0  \\ 
LLaMA3-8B-base                  & 42.5      & 10.0      & 36.0      & 53.7      & 20.0      & 47.0      & 37.5        & 0.0        & 30.0       \\
LLaMA3-8B-chat                  & 46.3      & 15.0      & 40.0      & 58.8      & 25.0      & 52.0      & 38.8        & 0.0        & 31.0       \\
LLaMA3.1-8B     & 56.3 & 10.0 & 47.0 & 46.3 & 10.0 & 39.0 & 43.8 & 0.0  & 35.0  \\ 
LLaMA3.1-70B    & 42.5 & 10.0 & 36.0 & 53.8 & 20.0 & 47.0 & 37.5 & 0.0  & 30.0  \\ 
LLaMA3.1-405B   & 48.8 & 5.0  & 40.0 & 41.3 & 5.0  & 34.0 & 47.5 & 10.0 & 40.0  \\ 
\hline
Gemma-7B        & 26.3 & 5.0  & 22.0 & 22.5 & 0.0  & 18.0 & 27.5 & 0.0  & 22.0  \\ 
Gemma2-9B       & 23.8 & 5.0  & 20.0 & 21.3 & 0.0  & 17.0 & 27.5 & 5.0  & 23.0  \\
\hline
GPT-3.5-turbo   & 51.3 & 0.0  & 41.0 & 53.8 & 15.0 & 46.0 & 32.5 & 15.0 & 29.0  \\ 
GPT-4o-mini     & 35.0 & 5.0  & 29.0 & 40.0 & 10.0 & 34.0 & 35.0 & 10.0 & 30.0  \\ 
\hline
ChatGLM3-6B-base                & 47.5      & 5.0       & 39.0      & \textbf{60.0}      & 25.0      & 53.0      & 50.0        & 5.0        & 41.0       \\
ChatGLM3-6B-chat                & 38.8      & 5.0       & 32.0      & 51.3      & 30.0      & 47.0      & 41.3        & 5.0        & 34.0       \\
ChatGLM4-9B     & 56.3 & 10.0 & 47.0 & \textbf{73.8} & 5.0  & \textbf{60.0} & \textbf{61.3} & 10.0 & 51.0  \\ 
\hline
Baichuan2-7B-base               & 50.0      & 5.0       & 41.0      & 47.5      & 0.0       & 38.0      & 38.8        & 15.0       & 34.0       \\
Baichuan2-7B-chat               & 36.3      & 5.0       & 30.0      & 51.3      & 25.0      & 46.0      & 40.0        & 0.0        & 32.0       \\
\hline
Qwen1.5-7B-base                 & 53.8      & 15.0      & 46.0      & \textbf{60.0}      & 10.0      & 50.0      & 53.8        & 10.0       & 45.0       \\
Qwen1.5-7B-chat                 & 47.5      & 15.0      & 41.0      & \textbf{63.8}      & 15.0      & 54.0      & 48.8        & 5.0        & 40.0       \\
Qwen1.5-110B    & \textbf{60.0} & 15.0 & 51.0 & \textbf{82.5} & 35.0 & \textbf{73.0} & \textbf{63.8} & 45.0 & \textbf{60.0}  \\ 
Qwen2-7B        & \textbf{66.3} & 15.0 & 56.0 & \textbf{70.0} & 25.0 & \textbf{61.0} & \textbf{65.0} & 10.0 & 54.0  \\ 
Qwen2-70B       & \textbf{70.0} & 30.0 & \textbf{62.0} & \textbf{77.5} & 45.0 & \textbf{71.0} & \textbf{68.8} & 45.0 & \textbf{64.0}  \\ 
\hline
Yi-6B-base                      & \textbf{61.3}      & 15.0      & 52.0      & \textbf{65.0}      & 5.0       & 53.0      & \textbf{60.0}        & 10.0       & 50.0       \\
Yi-6B-chat                      & \textbf{62.5}      & 0.0       & 50.0      & \textbf{70.0}      & 30.0      & \textbf{62.0}      & 56.3        & 5.0        & 46.0     \\
\hline
Cert. Bar (top 10\% human) & - & - & 60.0 & - & - & 60.0 & - & - & 60.0 \\
\bottomrule
\end{tabular}
\end{table*}

\subsection{Construction of UrbanPlanBench}
The motivation for UrbanPlanBench stems from the need to quantitatively assess the extent to which LLMs acquire expertise in urban planning. 
Specifically, we aim to answer the fundamental question of whether LLMs can match the proficiency of human planners, given that urban planning is inherently a human-centered field. 
To achieve this goal, we have constructed UrbanPlanBench based on the latest available real-world urban planning qualification exam in China, which serves as the standard for certifying registered urban planners. 
This benchmark evaluates LLMs from the following three critical perspectives (subjects) of urban planning: \begin{itemize}[leftmargin=*]
    \item \textbf{S1: Fundamental principles.} This subject delves into topics concerning cities and urban development, basic know-how of urban planning, urban land use and spatial layout, as well as practical implementation of urban planning.
    It reflects LLMs' grasp of the foundational theories underlying urban development and the discipline of urban planning.
    \item \textbf{S2: Professional knowledge.} This subject covers knowledge from eight professional fields that are closely related to urban planning, which include architecture, urban transportation, municipal public facilities, information technology application in urban planning, urban economics, urban geography, urban sociology, and urban ecology and environment. 
    It measures LLMs' proficiency, familiarity, and comprehension across various disciplines relevant to urban planning.
    \item \textbf{S3: Management and regulations.} Management covers urban planning formulation and approval management, implementation management, supervision and inspection, and professional ethics. 
    Regulations include foundational knowledge in administrative and urban planning laws, complementary regulations, technical standards and specifications, and other relevant laws and policies.
\end{itemize} 
By incorporating the above diverse perspectives, UrbanPlanBench forms a challenging testbed that comprehensively evaluates the mastery of urban planning skills for LLMs, shedding light on their capabilities in this complex domain.

In constructing UrbanPlanBench, we adopted the widely used multiple-choice question (MCQ) format~\citep{hendrycks2020measuring}, due to its efficacy in assessing LLMs' understanding and reasoning capabilities, with rigorously defined accuracy. 
For each of the three aforementioned perspectives, we crafted 100 MCQs. 
In each category, the initial 80 MCQs feature four choices, with only one correct answer (MCS-S).
To further challenge the LLMs, the remaining 20 MCQs in each perspective include five choices, with two to four correct options (MCQ-M). 
It is worth noting that MCQ-M questions are much more difficult than MCQ-S questions, as the accuracy of random guess drops from 25\% to 4\%.
All questions in UrbanPlanBench were curated from PDF or Microsoft Word documents, and meticulously transformed into a structured CSV format through careful parsing and annotation by the authors.
These questions were presented to the LLMs through prompts, as demonstrated in Figure \ref{fig::benchmark}.

In the context of the urban planner qualification exams, achieving a score of 60 out of 100 MCQs correctly answered across all subjects stands as a crucial criterion for certification.
These exams pose a significant challenge even for human participants, with only 10\% passing annually. 
Therefore, if LLMs can consistently answer 60 MCQs correctly across the three subjects, it suggests they have attained a level of urban planning expertise comparable to that of registered human planners, signifying the top 10\% of human-level proficiency.
By adhering to the rigorous standards set by real-world examinations, our constructed benchmark aims to offer tangible insights into the capabilities of LLMs that can be directly compared with those of human planners. 
Subsequently, we evaluate a diverse array of advanced LLMs on this benchmark to comprehensively scrutinize their urban planning abilities.

\subsection{Evaluation Results}

In our experimental evaluation, we prompted multiple advanced LLMs to respond to all questions presented in the introduced UrbanPlanBench. 
For each question, we selected the option with the highest output probability by each LLM as its final response~\citep{hendrycks2020measuring}, and then calculated the average accuracy within different subjects. 
Table \ref{tab::benchmark} illustrates the benchmarking results of LLMs, detailing the accuracy for both the 80 MCQ-S (single correct option) questions and the 20 MCQ-M (multiple correct options) questions separately, along with the accuracy for the entire set of 100 questions within each subject.

\subsubsection{Overall planning capabilities}
We have the following empirical findings:
\begin{itemize}[leftmargin=*]
    \item We observe that current advanced LLMs demonstrate a substantial level of proficiency in urban planning expertise. 
    Across all three subjects, the accuracy rates of all LLMs notably surpass random guess predictions, indicating the effectiveness of large-scale pretraining and supervised fine-tuning in equipping these models with urban planning memorization and reasoning abilities. 
    Specifically, the highest-performing LLM achieves approximately 2.98, 3.51, and 3.08 times higher accuracy than random guess predictions in fundamental principles, professional knowledge, and management and regulations, respectively. 
    Moreover, we observe that 9 LLMs achieve at least 50.0\% accuracy in at least one subject, underscoring their mastery of urban planning expertise.
    \item Despite these promising results, LLMs still lag significantly behind professional human planners in terms of performance. 
    All 25 evaluated LLMs, except for Qwen2-70B, fail to exceed the certification bar for professional human planners, \textit{i.e.} 60.0\% accuracy in all three subjects.
    Specifically, out of 75 cases comprising 25 different LLMs and 3 subjects, only 8 times does an LLM exceed the 60\% accuracy certification bar which roughly aligns with top 10\% human proficiency levels.
    This indicates that most of the LLMs evaluated in this study are not capable of passing the urban planning qualification exam, highlighting their insufficient urban planning capabilities compared to certified human planners. 
    \item We find that LLMs perform notably worse on MCQ-M questions compared to MCQ-S questions. 
    This discrepancy is understandable, given the increased complexity of MCQ-M questions, which feature a set of 25 potential answers, much larger than MCQ-S questions that only have 4 potential answers.
    Specifically, we observe zero accuracy in 15 out of 75 cases, indicating a performance level even below random guess predictions. 
    These findings suggest that, while most existing benchmarks for LLMs primarily focus on MCQ-S questions, it may be necessary to include more challenging MCQ-M tests to comprehensively evaluate the capabilities of LLMs in specialized domains such as urban planning.
    \item Surprisingly, we find that Qwen2-70B achieved accuracy of 62.0\%, 71.0\%, and 64.0\% on the three subjects, making it the first and only LLM to surpass the 60\% certification threshold of professional human planners. 
The inspiring results highlight the huge potential of LLMs to assist human planners in practical urban planning tasks.
\end{itemize}

\subsubsection{Subject imbalance}
The results in Table \ref{tab::benchmark} reveal an obvious imbalance in the performance of LLMs across the three distinct subjects evaluated in UrbanPlanBench. 
Specifically, we find that the average accuracy of the 25 LLMs on the three subjects is 38.24\%, 44.16\%, and 34.52\%, respectively. 
Particularly, LLMs demonstrate significantly better performance on S2 (professional knowledge) compared to the other two subjects, S1 (fundamental principles) and S3 (management and regulations). 
Moreover, 68\% LLMs achieve accuracy lower than 45.0\% in both S1 and S3, and only 16\% models achieves over 50.0\% accuracy in these two subjects. 

Upon closer examination of the definitions of the three subjects, we observe that S2 covers a broader range of general and diverse topics, potentially overlapping with the pretraining and SFT data of these LLMs. 
In contrast, S1 and S3 focus more on domain-specific contents, emphasizing specialized urban planning concepts that may be insufficiently represented in the training data. 
These findings underscore the need to develop a specialized SFT dataset tailored specifically to urban planning to enhance the performance of LLMs in this critical domain.

\subsubsection{Language bias}
UrbanPlanBench is a Chinese benchmark sourced from questions of urban planning exams in China, thus most of the evaluated LLMs are also Chinese LLMs which are pre-trained and finetuned with large-scale Chinese textual data.
Still, we include three English-primary LLM series for comparison, namely LLaMA, Gemma, and GPT.
The results highlight a notable difference between the performance of Chinese LLMs and two English-primary LLMs, particularly evident in S3 (management and regulations). 
The average accuracy of the three English-primary LLM series in S3 is 26.77\%, representing a significant 41.7\% relative gap compared to the other twelve Chinese LLMs, which exhibit an average accuracy of 45.92\%.

The disparity in performance between the English-primary LLMs and other Chinese LLMs is less pronounced in S1 and S2, with gaps of 8.9\% and 1.8\%, respectively. 
For example, the LLaMA3.1-8B and LLaMA3-8B-chat model surpass 7 and 4 Chinese LLMs in terms of accuracy in S1 and S2, respectively. 
These results suggest a potential difference in the adaptability of English-primary LLMs compared to their Chinese counterparts in comprehending and interpreting the specific regulations and management aspects inherent in urban planning contexts, emphasizing the importance of considering language-specific nuances in LLM performance evaluation and application.

\begin{figure}[t]
  \centering
  \includegraphics[width=\linewidth]{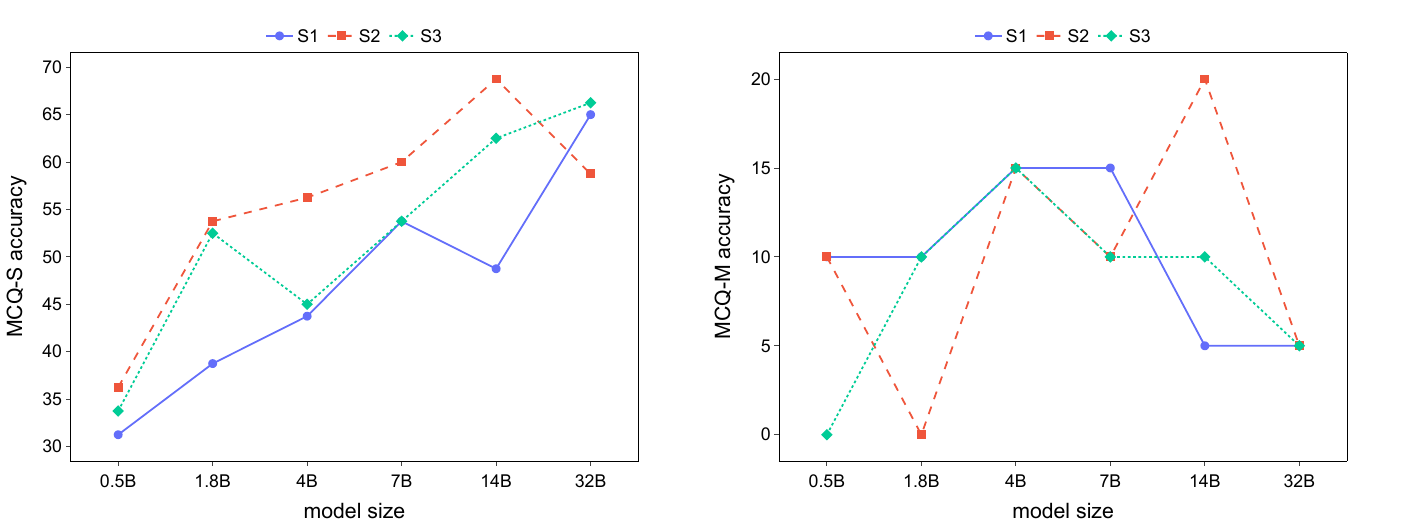}
  \vspace{-10px}
  \caption{Performance of different model sizes. The LLM model Qwen1.5 is adopted. MCS-S and MCQ-M indicate MCQs with one single correct answer and multiple correct answers, respectively. \textbf{(Left)} Accuracy on MCQ-S questions. \textbf{(Right)} Accuracy on MCQ-M questions.}
  \label{fig::scale_infer}
  \vspace{-15px}
\end{figure}

\subsubsection{Scaling effect}
Researchers have consistently observed scaling laws of neural language models, particularly LLMs, where scaling up models can lead to substantial performance improvements~\citep{kaplan2020scaling} and even emergent abilities~\citep{wei2022emergent}.
From the results we can observe that larger models generally achieve higher accuracy. 
For instance, LLaMA3.1-405B improved the performance by 53.8\%, 47.8\%, and 135.3\% on S1, S2, and S3, respectively, in comparison to LLaMA-7B.

To further investigate this phenomenon, we evaluated Qwen1.5 models of varying parameter scales, ranging from 0.8B to 32B parameters, on UrbanPlanBench. 
We calculated accuracy across three subjects, with MCQ-S and MCQ-M questions examined separately. 
The results, depicted in Figure \ref{fig::scale_infer}, showcase a notable scaling effect, particularly evident in MCQ-S questions, aligning with previous literature. 
Across all three subjects, we observed approximately a 100\% increase in accuracy for MCQ-S questions by scaling up models, with the largest improvement of 108.0\% seen in S1, followed by 96.3\% and 89.7\% improvements in S3 and S2, respectively. 
Remarkably, the Qwen1.5-14B and Qwen1.5-32B models achieved over 60\% accuracy in two of the three subjects, signaling their potential to rival professional human planners.

These findings underscore the validity of scaling laws in LLMs where larger models demonstrate enhanced understanding and reasoning capabilities, as evidenced in our specialized benchmark. 
However, we also observed that accuracy on MCQ-M questions remained low despite increasing model sizes. 
Given the increasing difficulty of MCQ-M tests, merely scaling up LLMs may prove insufficient, necessitating advanced techniques such as retrieval augmented generation (RAG)~\citep{lewis2020retrieval} to bolster the urban planning expertise of LLMs for addressing intricate MCQ-M questions.
We show the benefits of advanced inference-time prompting techniques in Section \ref{sec:prompting}.

\begin{table*}[t]
\centering
\caption{Accuracy (\%) on three subjects of UrbanPlanBench using RAG and CoT techniques. S and M indicate MCQs with one single correct answer and multiple correct answers, respectively. Full represents the overall accuracy of both types of MCQs.}
\label{tab::RAGCoT}
\vspace{-10px}
\begin{tabular}{c|ccc|ccc|ccc}
\toprule
\multirow{2}{*}{\textbf{Model}} & \multicolumn{3}{c|}{\textbf{S1}} & \multicolumn{3}{c|}{\textbf{S2}} & \multicolumn{3}{c}{\textbf{S3}} \\
                                & \textbf{S}   & \textbf{M}  & \textbf{Full}  & \textbf{S}   & \textbf{M}  & \textbf{Full}  & \textbf{S}    & \textbf{M}    & \textbf{Full}   \\
\midrule
GPT-4o-mini     & 40.0 & 5.0  & 33.0 & 45.0 & 5.0  & 37.0 & 42.5 & 15.0 & 37.0  \\ 
\hline
RAG\_direct      & 52.5 & 5.0  & 43.0 & 68.8 & 45.0 & 64.0 & 53.8 & 5.0  & 48.0  \\ 
RAG\_qa      & 58.8 & 5.0  & 48.0 & 70.0 & 30.0 & 62.0 & 56.3 & 20.0 & 49.0  \\ 
COT\_fs      & 60.0 & 15.0 & 51.0 & 70.0 & 35.0 & 63.0 & 58.8 & 25.0 & 53.0  \\ 
COT\_zs      & 53.8 & 10.0 & 45.0 & 72.5 & 35.0 & 65.0 & 51.3 & 35.0 & 48.0  \\ 
\hline
Cert. Bar (top 10\% human) & - & - & 60.0 & - & - & 60.0 & - & - & 60.0 \\
\bottomrule
\end{tabular}
\vspace{-10px}
\end{table*}

\subsubsection{Longitudinal studies}
To track the evolution of LLM performance over time, we compare different generations of LLMs.
We can observe that later generations of LLMs indeed achieve substantially better performance than their corrsponding earlier version in most cases. 
For example, LLaMA3.1-8B improves the accuracy on S1 by 105\% against LLaMA-13B, Qwen2-7B improves the accuracy on S2 by 17.3\% against Qwen1.5-7B, and ChatGLM4-9B improves the accuracy on S3 by 13.3\% against ChatGLM3-6B. 
The longitudinal studies confirm that the progress made in data quality, model structure, and training algorithm effectively enhances the ability of LLMs to understand and deal with complex urban planning problems. 
Nevertheless, the enhanced model capabilities alone are still not effective in dealing with MCQ-M questions, indicating the necessity of incorporating advanced techniques such as RAG~\citep{lewis2020retrieval} and COT~\citep{wei2022chain}.

\subsection{Prompting Techniques}\label{sec:prompting}

Prompting techniques can significantly enhance LLMs in answering complicated questions and improving their reasoning capabilities.
Here in UrbanPlanBench, we use GPT-4o-mini as the base model to validate the effectiveness of RAG~\cite{lewis2020retrieval} and CoT~\cite{wei2022chain} for knowledge augmentation of LLMs, as well as to evaluate their enhancement of LLMs' competence in the field of urban planning. 
Table \ref{tab::RAGCoT} illustrates the results of different prompting techniques.
Specifically, we utilize Self-RAG~\cite{asai2023self} to retrieve matches in RAG experiments, where RAG\_direct denotes that the relevant textbooks and previous years' questions are directly used as the content of the knowledge base, and RAG\_qa denotes that these contents are first processed into high-quality QA. 
With respect to COT, we adopt both few-shot-CoT and zero-shot-CoT, denoted as CoT\_fs and CoT\_zs, respectively. 
From the results, we have the following observations:

\begin{itemize}[leftmargin=*]
\item \textit{RAG prompting.} The introduction of RAG significantly improves the performance on each subject.
Specifically, RAG\_direct and RAG\_qa improve the accuracy on S1 by 30.3\% and 45.5\%, respectively, compared to the GPT-4o-mini base model. 
In S2, RAG\_direct and RAG\_qa achieve an overall average accuracy of 64\% and 62\%, respectively, and both MCQ-S and MCQ-M are improved by more than 52.9\%, reaching the level of professional urban planners. 
In MCQ-M of S3, the GPT-4o-mini base model performs better than RAG\_direct, but the accurate information retrieval still ensures a high accuracy rate in MCQ-S for RAG models with about 32.5\% improvements against the base model.

\item \textit{CoT prompting.} The introduction of CoT technology leads to stronger reasoning ability of LLMs and significantly improves the performance of each subject. 
Specifically, CoT\_fs and CoT\_zs improved the accuracy by 54.5\% and 36.4\% in S1, 70.3\% and 75.7\% in S2, and 43.2\% and 29.7\% in S3, respectively.
Notably, COT substantially improves the performance of LLMs in answering MCQ-M questions.
For example, COT\_zs increases the accuracy on MCQ-M by 100\%, 600\%, and 133\% in the three subjects.
As there exist more than one correct options in MCQ-M questions which are much more complicated than MCQ-S ones, the above results confirm the effectiveness of COT in boosting the reasoning ability of LLMs.

The above benchmarking results illustrate the large potential of LLMs in urban planning.
In practical urban planning scenarios, LLMs can be smoothly integrated into planners' workflow using appropriate prompts, and we provide two example cases of urban planning text polishment and proofreading in Appendix \ref{app:case}.

\end{itemize}

\section{Fine-tuning LLMs with UrbanPlanText}

The inherent knowledge of LLMs proves insufficient when confronted with specialized urban planning queries, highlighting a deficiency in domain-specific understanding. 
SFT emerges as a widely adopted technique for tailoring LLMs towards specific domains by infusing them with related knowledge and data. 
Notably, SFT datasets for LLMs typically consist of sample pairs comprising instructions and corresponding responses. 
However, existing urban planning knowledge is scattered across unannotated textual resources, presenting a challenge in sourcing relevant data for SFT. 
Towards this end, we initially gathered materials from seven urban planning textbooks, along with archives of urban planning exams spanning the past eight years. 
Subsequently, we derive instruction pairs from these materials, leading to the largest-ever SFT dataset tailored for urban planning. 

\subsection{Dataset Construction}

\noindent\textbf{Data Sources.}
Our urban planning textual data collection primarily focuses on two key sources: urban planning textbooks and past urban planning exams, with the overview of the dataset's statistics shown in Table \ref{tab::data}. 
Urban planning textbooks encompass a wide spectrum of general knowledge about urban planning, thus leveraging data from textbooks enables LLMs to establish a foundational understanding of the specific domain, mirroring the approach taken by human planners who frequently refer to textbooks as their primary learning and training resources.

In addition, questions found in urban planning exams adhere to specific formats and emphasize particular areas. 
Therefore, fine-tuning LLMs with data extracted from real exams serves to further refine their abilities, enhancing their accuracy in addressing domain-specific exam questions. 
Particularly, as the introduced benchmark UrbanPlanBench is sourced from the latest publicly available urban planning exam in China held in 2022, we utilize exam questions predating 2022 for the collection of SFT data, spanning eight years.

\begin{table}[t]
\small
\centering
\caption{Statistics of different sources for UrbanPlanText.}
\vspace{-10px}
\label{tab::data}
\begin{tabular}{cccc}
\toprule
\textbf{Category}                    & \textbf{Name} & \textbf{\#Words} & \textbf{\#Samples} \\
\midrule

\multirow{2}{*}{Past Exams} &  MCQ    & 619,810  & 2,397 \\
                            &  dialog    & 350,080  & 4,139 \\
\hline
\multirow{12}{*}{Textbooks}  
                            & \makecell{\textit{Principles of} \\ \textit{urban planning}}  
                            & 470,621          & 5,091           \\
                            \cline{2-4}
                            & \makecell{\textit{Knowledge of} \\ \textit{urban planning}}     
                            & 457,236         & 9,307           \\
                            \cline{2-4}
                            & \makecell{\textit{Urban planning management} \\ \textit{and regulations}}    
                            & 313,246         & 4,347           \\
                            \cline{2-4}
                            & \makecell{\textit{Urban planning practice}}     
                            & 120,155          & 3,589           \\
                            \cline{2-4}
                            & \makecell{\textit{Detailed regulatory plan}}     
                            & 174,626        & 246           \\
                            \cline{2-4}
                            & \makecell{\textit{History of urban} \\ \textit{construction in China}} 
                            & 156,923        & 608           \\
                            \cline{2-4}
                            & \makecell{\textit{Additional contents of} \\ \textit{urban planning exams}}    
                            & 60,371         & 1,610           \\
\hline
& Total                & 2,723,068         & 31,334   \\
\bottomrule
\end{tabular}
\vspace{-10px}
\end{table}

\begin{figure}[t]
  \centering
  \includegraphics[width=\linewidth]{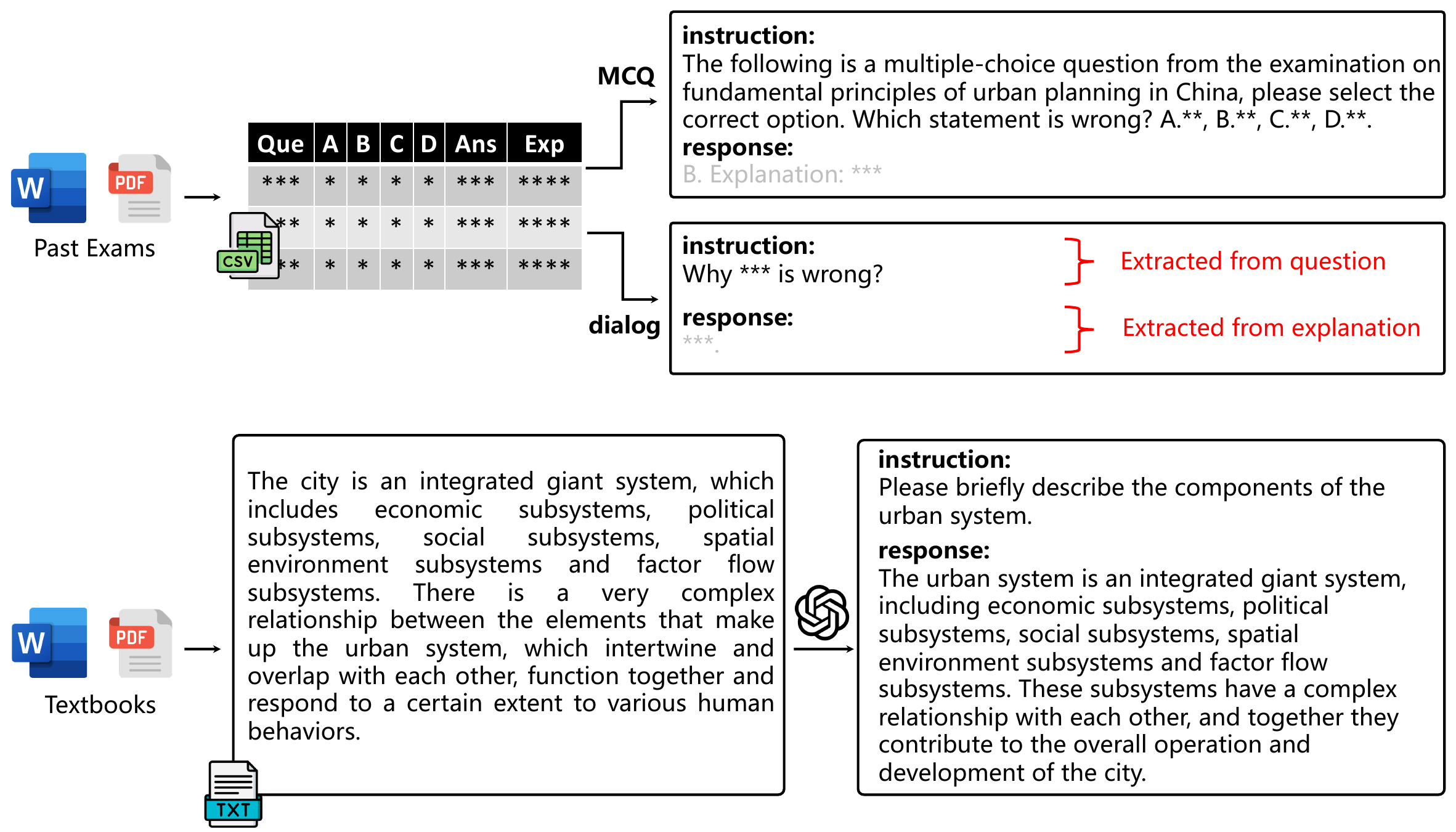}
\vspace{-20px}
  \caption{Data collection and process of UrbanPlanText.
  Past exam questions are first annotated by the authors into structured CSV files and then transformed into MCQ-type instruction pairs and dialog-style instruction pairs.
  Textbooks are first parsed into textual files, from which instruction pairs are generated automatically by prompting OpenAI's ChatGPT model. Contents are translated from Chinese to English.}
  \label{fig::data}
\vspace{-20px}
\end{figure}

\noindent\textbf{Data Processing.}
The collected original materials encompass a variety of formats, predominantly stored as PDF or Microsoft Word documents. 
To facilitate further processing, these materials are first parsed into plain text format. 
In the case of urban planning exams from previous years, an additional step is taken to transform these MCQs into a structured CSV format before they are processed into instruction pairs. 
These instruction pairs are designed to include the system prompt, the question itself, and the provided options, while the response comprises the correct answer accompanied by explanations.
Meanwhile, dialog-style instruction pairs are derived from MCQs to further enrich the training materials, as depicted in Figure \ref{fig::data}. 

For urban planning textbooks, which primarily contain descriptive text without readily available instruction pairs, more data process is needed. 
Here, we employ OpenAI's ChatGPT to automatically generate instruction pairs from the descriptive text by prompting, as demonstrated in Figure \ref{fig::data}. 
This process ensures the conversion of all collected materials into a standardized format suitable for subsequent SFT.
The prompt template of data generation is provided in Appendix \ref{app:prompt}.

\subsection{SFT Data Quality}

In our experiments, we tried different approaches to generate instruction pairs including OpenAI's ChatGPT, Ernie, and the opensourced Bonito framework. 
Eventually we employed OpenAI's ChatGPT due to its better performance.
It is worthwhile to notice that using LLMs to generate training data has become a common practice proven to be effective and widely adopted by related literature~\cite{betker2023improving,esser2024scaling}. To verify the data quality, we invited five domain experts with graduated degree in urban planning to judge the generated instruction pairs, comparing different approaches. Specifically, we sample 100 instruction pairs generated by different approaches, and asked the domain experts to score the generated data from the perspective of both correctiveness and informativeness. The results are demonstrated in Table \ref{tab:expert_evaluation}.

\subsection{SFT Results}

\begin{table}[t]
\centering
\caption{Correctiveness and informativeness of the generated SFT data by human experts.}
\label{tab:expert_evaluation}
\vspace{-10px}
\begin{tabular}{c|c|c|c|c}
\toprule
\multirow{2}{*}{\textbf{Expert}} & \multirow{2}{*}{\textbf{Metric}} & \multicolumn{3}{c}{\textbf{Model}} \\ 
                        &                         & \textbf{ChatGPT} & \textbf{Ernie} & \textbf{Bonito} \\ 
\midrule
\multirow{2}{*}{1}      & Correctiveness           & 9.6     & 9.6   & 7.2    \\ 
                        & Informativeness          & 9.4     & 8.8   & 5.8    \\ \hline
\multirow{2}{*}{2}      & Correctiveness           & 9.2     & 7.4   & 7.6    \\ 
                        & Informativeness          & 8.4     & 6.8   & 7.6    \\ \hline
\multirow{2}{*}{3}      & Correctiveness           & 6.2     & 6.0   & 5.2    \\  
                        & Informativeness          & 6.2     & 6.2   & 4.6    \\ \hline
\multirow{2}{*}{4}      & Correctiveness           & 8.8     & 8.0   & 4.8    \\  
                        & Informativeness          & 8.6     & 7.4   & 4.2    \\ \hline
\multirow{2}{*}{5}      & Correctiveness           & 7.6     & 7.4   & 6.4    \\ 
                        & Informativeness          & 7.6     & 7.6   & 5.2    \\ 
\bottomrule
\end{tabular}
\end{table}

\begin{table}[t]
\small
\centering
\caption{Accuracy (\%) of LLMs on three subjects of UrbanPlanBench after SFT on UrbanPlanText. S and M indicate MCQs with one single correct answer and multiple correct answers, respectively. Full represents the overall accuracy of both types of MCQs. \textbf{Bold numbers} indicate that the performance improves against the corresponding pre-SFT model.}
\label{tab::sft}
\vspace{-10px}
\begin{tabular}{c@{\hspace{3pt}}|c@{\hspace{3pt}}c@{\hspace{3pt}}c|c@{\hspace{3pt}}c@{\hspace{3pt}}c|c@{\hspace{3pt}}c@{\hspace{3pt}}c}

\toprule
\multirow{2}{*}{\textbf{Model}} & \multicolumn{3}{c|}{\textbf{S1}} & \multicolumn{3}{c|}{\textbf{S2}} & \multicolumn{3}{c}{\textbf{S3}}  \\
                                & \textbf{S}   & \textbf{M}  & \textbf{Full}  & \textbf{S}   & \textbf{M}  & \textbf{Full}  & \textbf{S}    & \textbf{M}    & \textbf{Full} \\
\midrule
LLaMA3-8B-base                  & 40.0      &   5.0    &  33.0     & 57.5      &  25.0     &  \textbf{51.0}     & 41.3        & 0.0        &  \textbf{33.0}  \\
LLaMA3-8B-chat                  & 42.5      &  10.0     &  36.0     &  51.3     &  10.0     & 43.0      &   33.8      &  5.0      &  28.0    \\
ChatGLM3-6B-base                & 50.0      &   5.0    &  \textbf{41.0}     &  58.8     &  35.0     &  \textbf{54.0}     &   55.0      &   5.0     &   \textbf{45.0}  \\
ChatGLM3-6B-chat                & 35.0      &  5.0     &  29.0     &  52.5     &  25.0     & 47.0      &   42.5      &  5.0      &  \textbf{35.0}    \\
Baichuan2-7B-base               & 46.3      &  5.0     &  38.0     &  56.3     &  5.0     &  \textbf{46.0}     &   43.8      &   15.0     &  \textbf{38.0}    \\
Baichuan2-7B-chat               & 43.8      &   5.0    &  \textbf{36.0}     &  46.3     &  20.0     &  41.0     &    31.3     &  0.0      &  25.0    \\
Qwen1.5-7B-base                 & 50.0      &   10.0    & 42.0      &  63.8     &  5.0     &  \textbf{52.0}     &   53.8     &  10.0      & 45.0    \\
Qwen1.5-7B-chat                 & 48.8      &  15.0     & \textbf{42.0}      &  63.8     &  5.0     &  52.0     &   50.0      &  10.0      &  \textbf{42.0}    \\
Yi-6B-base                      & 58.8      &  15.0     & 50.0      & 68.8      &  0.0     & \textbf{55.0}      &  61.3       &   0.0     &   49.0     \\
Yi-6B-chat                      & 62.5      &  0.0     &  50.0     &  66.3     &  35.0     &  60.0     &  63.8       &   0.0     &  \textbf{51.0}     \\
\bottomrule
\end{tabular}
\vspace{-10px}
\end{table}

We leveraged our constructed UrbanPlanText dataset to fine-tune LLMs using LLaMA-Factory~\citep{zheng2024llamafactory}. 
Employing lora~\citep{hu2021lora} to accelerate the SFT process, we fine-tuned all models for three epochs on one single Nvidia A100 GPU, which takes about 4 hours. 
Subsequently, we evaluated the fine-tuned LLMs again on UrbanPlanBench, with the results detailed in Table \ref{tab::sft}. 
Notably, we observed a significant enhancement in performance, particularly in S3 (management and regulations). 
Specifically, 60\% of LLMs exhibited improved accuracy on full questions of S3, and 70\% demonstrated enhanced accuracy on the MCQ-S questions of S3, with the average accuracy of LLMs improved by 2.1\% compared to their pre-SFT counterparts. 
Given that S3 was previously the weakest subject for LLMs according to Table \ref{tab::benchmark}, these findings underscore the effectiveness of enhancing the domain-specific capabilities of LLMs through SFT. 
Additionally, for the previously strongest subject, S2 (professional knowledge), LLMs maintained competitive performance, with an average accuracy of 50.1\%, similar to the pre-SFT average accuracy of 50.2\%.

\begin{figure}[t]
  \centering
  \includegraphics[width=\linewidth]{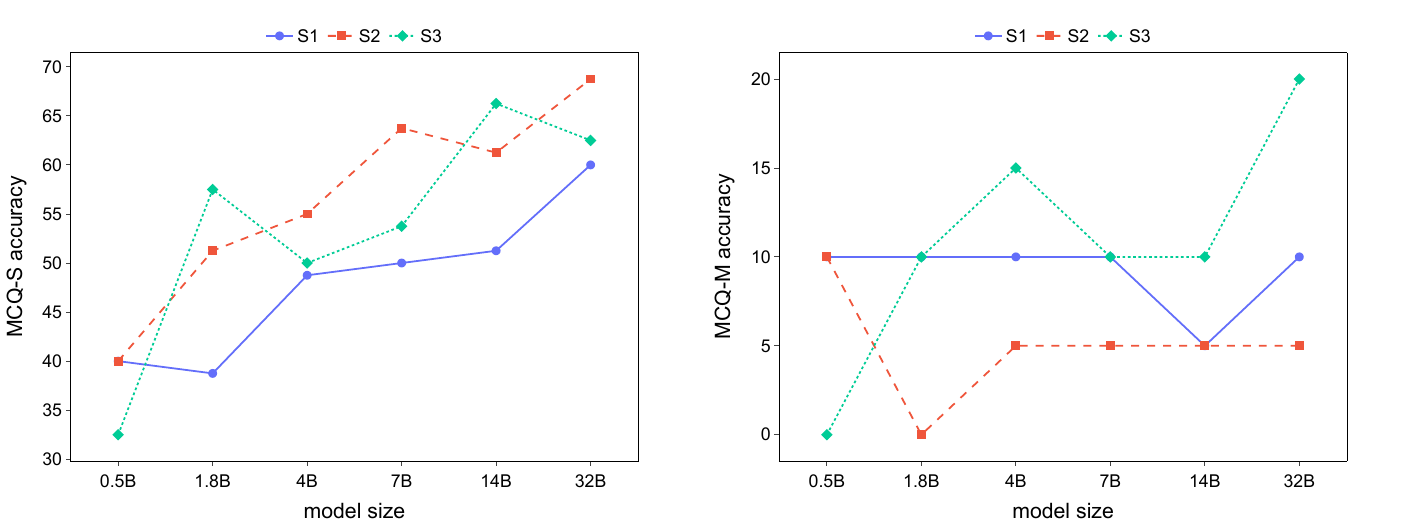}
    \vspace{-15px}
  \caption{Performance of different model sizes after SFT on UrbanPlanText. The LLM model Qwen1.5 is adopted. MCS-S and MCQ-M indicate MCQs with one single correct answer and multiple correct answers, respectively. \textbf{(Left)} Accuracy on MCQ-S questions. \textbf{(Right)} Accuracy on MCQ-M questions.}
  \label{fig::scale_sft}
  \vspace{-10px}
\end{figure}

Additionally, we conducted SFT experiments on LLMs of varying sizes using UrbanPlanText and subsequently evaluated their performance on UrbanPlanBench. 
Aligning with previous benchmarking experiments, we still employed the Qwen1.5 model across a spectrum of parameter sizes ranging from 0.8B to 32B.
We illustrated their post-SFT performance in Figure \ref{fig::scale_sft}. 
Similar to previous observations, we can observe a clear scaling effect on the accuracy of MCQ-S questions, with larger models demonstrating substantially improved performance compared to their smaller counterparts. 
Particularly, we noted that SFT yielded more substantial benefits for smaller models. 
For instance, the average accuracy on MCQ-S questions across all three subjects increased by 11.1\%, 1.7\%, and 6.0\% for the smallest three models (0.5B, 1.8B, and 4B) compared to previous results in Figure \ref{fig::scale_infer}, respectively, while the improvement was only 0.7\% for the largest 32B model. 
These findings hold significant practical implications, particularly as smaller models are more accessible to a broader user base at a considerably lower cost.

Similar to previous findings, the accuracy on MCQ-M questions does not improve with growing model sizes, which again confirming the inherent difficulty of MCQ-M questions.
The results in Figure \ref{fig::scale_sft}, in comparison to Section \ref{sec:prompting}, suggest that combining SFT and advanced prompting techniques can enhance performance on challenging questions and lead to better overall accuracy.

\section{Related Work}

\noindent\textbf{AI for urban planning.}
AI applications in urban planning offer promising solutions to the challenges posed by rapid urbanization, aiming to alleviate the burden on human planners~\citep{zheng2024survey}.
Current research predominantly focuses on urban design, generating layouts of various urban functionalities such as land use~\citep{zheng2023spatial}, transportation networks~\citep{zheng2023road,su2024metrognn}, buildings~\citep{qin2024text2city}, and points of interest (POIs)~\citep{wang2020reimagining}. 
These endeavors approach urban design as either a generation problem, utilizing existing urban data and generative models like diffusion models~\citep{qin2024text2city} and generative adversarial networks (GANs)~\citep{wang2020reimagining}, or as an optimization problem tackled through methods such as reinforcement learning (RL)~\citep{zheng2023spatial,zheng2023road,su2024metrognn,su2024large} to find more efficient layouts. 
Despite urban design, urban planners still devote significant time to handling urban planning-related text.
The emergence of LLMs has led to the development of specialized models tailored for urban planning tasks~\citep{wang2024transgpt,zhang2024trafficgpt,zhu2024plangpt}. 
For instance, TransGPT~\citep{wang2024transgpt} fine-tunes LLMs with large-scale transportation text to assist in transportation planning, while PlanGPT~\citep{zhu2024plangpt} equips LLMs with external knowledge and web search capabilities for various text-related tasks in urban planning.
However, these efforts often rely on case studies to demonstrate the effectiveness of LLMs in urban planning, underscoring the urgent need for a comprehensive benchmark to quantitatively assess the extent to which LLMs masters urban planning knowledge.

\noindent\textbf{Domain-specific benchmarks for LLMs.}
Benchmarks play a pivotal role in shaping the trajectory of AI research, serving as foundational tools that drive progress within the field~\citep{patterson2012better}. 
LLMs have demonstrated exceptional understanding, reasoning, and memorization abilities, as evidenced by benchmarks such as SuperGLUE~\citep{wang2019superglue}, BIG-Bench~\citep{srivastava2022beyond}, MMLU~\citep{hendrycks2020measuring}, and HELM~\citep{liang2022holistic}, which cover diverse Natural Language Processing (NLP) tasks. 
While general-purpose NLP benchmarks have provided valuable insights into LLM capabilities, domain-specific benchmarks are indispensable to understand LLMs' specialized expertise~\citep{chen2023bioinfo,koncel2023bizbench,fei2023lawbench,liu2024mathbench}. 
Examples include LawBench~\citep{fei2023lawbench}, which evaluates LLMs' legal capabilities in memorization, understanding, and application of legal knowledge, and BizBench~\citep{koncel2023bizbench}, which assesses LLMs' ability to reason about financial problems and synthesize code to accomplish Q\&A tasks over financial data. 
Additionally, MathBench~\citep{liu2024mathbench} evaluates LLMs' mathematical proficiency in answering theoretical questions and solving application problems. 
However, within the realm of urban planning, there is a notable absence of publicly available benchmarks, impeding the effective utilization of LLMs in this critical domain. 
In response to this gap, this paper proposes the first urban planning benchmark for LLMs, aiming to comprehensively evaluate their capabilities and guide technological advancements in this field.

\section{Conclusion and Future Work}

This paper introduces UrbanPlanBench and UrbanPlanText, the first urban planning benchmark and the largest-ever SFT dataset tailored for LLMs. 
These resources, along with open-sourced toolsets, provide comprehensive support for fine-tuning and evaluating LLMs in the critical domain of urban planning. 
Through a series of experiments involving multiple advanced LLMs, we have showcased their remarkable capabilities in mastering urban planning knowledge. 
However, there remains substantial untapped potential to fully leverage LLMs to enhance the productivity of human practitioners in this field. 
We envision that our findings will foster interdisciplinary collaboration between human planners and AI practitioners, paving the way for further exploration and the application of LLMs in influential real-world urban planning scenarios. 
Moving forward, our future work includes expanding both UrbanPlanBench and UrbanPlanText to incorporate multi-linguistic urban planning materials, thereby enabling broader use cases of the benchmark and dataset. 
Additionally, we aim to extend UrbanPlanBench into a multi-modal benchmark, integrating both imagery of urban plans and their corresponding descriptive text, further enriching the evaluation capabilities of LLMs in urban planning contexts.

\bibliographystyle{ACM-Reference-Format}
\bibliography{sample-base}


\begin{thebibliography}{47}


\ifx \showCODEN    \undefined \def \showCODEN     #1{\unskip}     \fi
\ifx \showISBNx    \undefined \def \showISBNx     #1{\unskip}     \fi
\ifx \showISBNxiii \undefined \def \showISBNxiii  #1{\unskip}     \fi
\ifx \showISSN     \undefined \def \showISSN      #1{\unskip}     \fi
\ifx \showLCCN     \undefined \def \showLCCN      #1{\unskip}     \fi
\ifx \shownote     \undefined \def \shownote      #1{#1}          \fi
\ifx \showarticletitle \undefined \def \showarticletitle #1{#1}   \fi
\ifx \showURL      \undefined \def \showURL       {\relax}        \fi
\providecommand\bibfield[2]{#2}
\providecommand\bibinfo[2]{#2}
\providecommand\natexlab[1]{#1}
\providecommand\showeprint[2][]{arXiv:#2}

\bibitem[Asai et~al\mbox{.}(2023)]%
        {asai2023self}
\bibfield{author}{\bibinfo{person}{Akari Asai}, \bibinfo{person}{Zeqiu Wu}, \bibinfo{person}{Yizhong Wang}, \bibinfo{person}{Avirup Sil}, {and} \bibinfo{person}{Hannaneh Hajishirzi}.} \bibinfo{year}{2023}\natexlab{}.
\newblock \showarticletitle{Self-rag: Learning to retrieve, generate, and critique through self-reflection}.
\newblock \bibinfo{journal}{\emph{arXiv preprint arXiv:2310.11511}} (\bibinfo{year}{2023}).
\newblock


\bibitem[Bai et~al\mbox{.}(2023)]%
        {qwen}
\bibfield{author}{\bibinfo{person}{Jinze Bai}, \bibinfo{person}{Shuai Bai}, \bibinfo{person}{Yunfei Chu}, \bibinfo{person}{Zeyu Cui}, \bibinfo{person}{Kai Dang}, \bibinfo{person}{Xiaodong Deng}, \bibinfo{person}{Yang Fan}, \bibinfo{person}{Wenbin Ge}, \bibinfo{person}{Yu Han}, \bibinfo{person}{Fei Huang}, \bibinfo{person}{Binyuan Hui}, \bibinfo{person}{Luo Ji}, \bibinfo{person}{Mei Li}, \bibinfo{person}{Junyang Lin}, \bibinfo{person}{Runji Lin}, \bibinfo{person}{Dayiheng Liu}, \bibinfo{person}{Gao Liu}, \bibinfo{person}{Chengqiang Lu}, \bibinfo{person}{Keming Lu}, \bibinfo{person}{Jianxin Ma}, \bibinfo{person}{Rui Men}, \bibinfo{person}{Xingzhang Ren}, \bibinfo{person}{Xuancheng Ren}, \bibinfo{person}{Chuanqi Tan}, \bibinfo{person}{Sinan Tan}, \bibinfo{person}{Jianhong Tu}, \bibinfo{person}{Peng Wang}, \bibinfo{person}{Shijie Wang}, \bibinfo{person}{Wei Wang}, \bibinfo{person}{Shengguang Wu}, \bibinfo{person}{Benfeng Xu}, \bibinfo{person}{Jin Xu}, \bibinfo{person}{An Yang}, \bibinfo{person}{Hao Yang},
  \bibinfo{person}{Jian Yang}, \bibinfo{person}{Shusheng Yang}, \bibinfo{person}{Yang Yao}, \bibinfo{person}{Bowen Yu}, \bibinfo{person}{Hongyi Yuan}, \bibinfo{person}{Zheng Yuan}, \bibinfo{person}{Jianwei Zhang}, \bibinfo{person}{Xingxuan Zhang}, \bibinfo{person}{Yichang Zhang}, \bibinfo{person}{Zhenru Zhang}, \bibinfo{person}{Chang Zhou}, \bibinfo{person}{Jingren Zhou}, \bibinfo{person}{Xiaohuan Zhou}, {and} \bibinfo{person}{Tianhang Zhu}.} \bibinfo{year}{2023}\natexlab{}.
\newblock \showarticletitle{Qwen Technical Report}.
\newblock \bibinfo{journal}{\emph{arXiv preprint arXiv:2309.16609}} (\bibinfo{year}{2023}).
\newblock


\bibitem[Baichuan(2023)]%
        {baichuan2023baichuan2}
\bibfield{author}{\bibinfo{person}{Baichuan}.} \bibinfo{year}{2023}\natexlab{}.
\newblock \showarticletitle{Baichuan 2: Open Large-scale Language Models}.
\newblock \bibinfo{journal}{\emph{arXiv preprint arXiv:2309.10305}} (\bibinfo{year}{2023}).
\newblock
\urldef\tempurl%
\url{https://arxiv.org/abs/2309.10305}
\showURL{%
\tempurl}


\bibitem[Betker et~al\mbox{.}(2023)]%
        {betker2023improving}
\bibfield{author}{\bibinfo{person}{James Betker}, \bibinfo{person}{Gabriel Goh}, \bibinfo{person}{Li Jing}, \bibinfo{person}{Tim Brooks}, \bibinfo{person}{Jianfeng Wang}, \bibinfo{person}{Linjie Li}, \bibinfo{person}{Long Ouyang}, \bibinfo{person}{Juntang Zhuang}, \bibinfo{person}{Joyce Lee}, \bibinfo{person}{Yufei Guo}, {et~al\mbox{.}}} \bibinfo{year}{2023}\natexlab{}.
\newblock \showarticletitle{Improving image generation with better captions}.
\newblock \bibinfo{journal}{\emph{Computer Science. https://cdn. openai. com/papers/dall-e-3. pdf}} \bibinfo{volume}{2}, \bibinfo{number}{3} (\bibinfo{year}{2023}), \bibinfo{pages}{8}.
\newblock


\bibitem[Chen et~al\mbox{.}(2021)]%
        {chen2021evaluating}
\bibfield{author}{\bibinfo{person}{Mark Chen}, \bibinfo{person}{Jerry Tworek}, \bibinfo{person}{Heewoo Jun}, \bibinfo{person}{Qiming Yuan}, \bibinfo{person}{Henrique Ponde de~Oliveira Pinto}, \bibinfo{person}{Jared Kaplan}, \bibinfo{person}{Harri Edwards}, \bibinfo{person}{Yuri Burda}, \bibinfo{person}{Nicholas Joseph}, \bibinfo{person}{Greg Brockman}, {et~al\mbox{.}}} \bibinfo{year}{2021}\natexlab{}.
\newblock \showarticletitle{Evaluating large language models trained on code}.
\newblock \bibinfo{journal}{\emph{arXiv preprint arXiv:2107.03374}} (\bibinfo{year}{2021}).
\newblock


\bibitem[Chen and Deng(2023)]%
        {chen2023bioinfo}
\bibfield{author}{\bibinfo{person}{Qiyuan Chen} {and} \bibinfo{person}{Cheng Deng}.} \bibinfo{year}{2023}\natexlab{}.
\newblock \showarticletitle{Bioinfo-Bench: A Simple Benchmark Framework for LLM Bioinformatics Skills Evaluation}.
\newblock \bibinfo{journal}{\emph{bioRxiv}} (\bibinfo{year}{2023}), \bibinfo{pages}{2023--10}.
\newblock


\bibitem[Esser et~al\mbox{.}(2024)]%
        {esser2024scaling}
\bibfield{author}{\bibinfo{person}{Patrick Esser}, \bibinfo{person}{Sumith Kulal}, \bibinfo{person}{Andreas Blattmann}, \bibinfo{person}{Rahim Entezari}, \bibinfo{person}{Jonas M{\"u}ller}, \bibinfo{person}{Harry Saini}, \bibinfo{person}{Yam Levi}, \bibinfo{person}{Dominik Lorenz}, \bibinfo{person}{Axel Sauer}, \bibinfo{person}{Frederic Boesel}, {et~al\mbox{.}}} \bibinfo{year}{2024}\natexlab{}.
\newblock \showarticletitle{Scaling rectified flow transformers for high-resolution image synthesis}. In \bibinfo{booktitle}{\emph{Forty-first International Conference on Machine Learning}}.
\newblock


\bibitem[Fei et~al\mbox{.}(2023)]%
        {fei2023lawbench}
\bibfield{author}{\bibinfo{person}{Zhiwei Fei}, \bibinfo{person}{Xiaoyu Shen}, \bibinfo{person}{Dawei Zhu}, \bibinfo{person}{Fengzhe Zhou}, \bibinfo{person}{Zhuo Han}, \bibinfo{person}{Songyang Zhang}, \bibinfo{person}{Kai Chen}, \bibinfo{person}{Zongwen Shen}, {and} \bibinfo{person}{Jidong Ge}.} \bibinfo{year}{2023}\natexlab{}.
\newblock \showarticletitle{Lawbench: Benchmarking legal knowledge of large language models}.
\newblock \bibinfo{journal}{\emph{arXiv preprint arXiv:2309.16289}} (\bibinfo{year}{2023}).
\newblock


\bibitem[Fu et~al\mbox{.}(2024)]%
        {fu2024towards}
\bibfield{author}{\bibinfo{person}{Jiayi Fu}, \bibinfo{person}{Haoying Han}, \bibinfo{person}{Xing Su}, {and} \bibinfo{person}{Chao Fan}.} \bibinfo{year}{2024}\natexlab{}.
\newblock \showarticletitle{Towards human-AI collaborative urban science research enabled by pre-trained large language models}.
\newblock \bibinfo{journal}{\emph{Urban Informatics}} \bibinfo{volume}{3}, \bibinfo{number}{1} (\bibinfo{year}{2024}), \bibinfo{pages}{8}.
\newblock


\bibitem[Gao et~al\mbox{.}(2023)]%
        {gao2023retrieval}
\bibfield{author}{\bibinfo{person}{Yunfan Gao}, \bibinfo{person}{Yun Xiong}, \bibinfo{person}{Xinyu Gao}, \bibinfo{person}{Kangxiang Jia}, \bibinfo{person}{Jinliu Pan}, \bibinfo{person}{Yuxi Bi}, \bibinfo{person}{Yi Dai}, \bibinfo{person}{Jiawei Sun}, {and} \bibinfo{person}{Haofen Wang}.} \bibinfo{year}{2023}\natexlab{}.
\newblock \showarticletitle{Retrieval-augmented generation for large language models: A survey}.
\newblock \bibinfo{journal}{\emph{arXiv preprint arXiv:2312.10997}} (\bibinfo{year}{2023}).
\newblock


\bibitem[GLM et~al\mbox{.}(2024)]%
        {glm2024chatglm}
\bibfield{author}{\bibinfo{person}{Team GLM}, \bibinfo{person}{Aohan Zeng}, \bibinfo{person}{Bin Xu}, \bibinfo{person}{Bowen Wang}, \bibinfo{person}{Chenhui Zhang}, \bibinfo{person}{Da Yin}, \bibinfo{person}{Diego Rojas}, \bibinfo{person}{Guanyu Feng}, \bibinfo{person}{Hanlin Zhao}, \bibinfo{person}{Hanyu Lai}, {et~al\mbox{.}}} \bibinfo{year}{2024}\natexlab{}.
\newblock \showarticletitle{ChatGLM: A Family of Large Language Models from GLM-130B to GLM-4 All Tools}.
\newblock \bibinfo{journal}{\emph{arXiv preprint arXiv:2406.12793}} (\bibinfo{year}{2024}).
\newblock


\bibitem[Hendrycks et~al\mbox{.}(2020)]%
        {hendrycks2020measuring}
\bibfield{author}{\bibinfo{person}{Dan Hendrycks}, \bibinfo{person}{Collin Burns}, \bibinfo{person}{Steven Basart}, \bibinfo{person}{Andy Zou}, \bibinfo{person}{Mantas Mazeika}, \bibinfo{person}{Dawn Song}, {and} \bibinfo{person}{Jacob Steinhardt}.} \bibinfo{year}{2020}\natexlab{}.
\newblock \showarticletitle{Measuring Massive Multitask Language Understanding}. In \bibinfo{booktitle}{\emph{International Conference on Learning Representations}}.
\newblock


\bibitem[Hu et~al\mbox{.}(2021)]%
        {hu2021lora}
\bibfield{author}{\bibinfo{person}{Edward~J Hu}, \bibinfo{person}{Yelong Shen}, \bibinfo{person}{Phillip Wallis}, \bibinfo{person}{Zeyuan Allen-Zhu}, \bibinfo{person}{Yuanzhi Li}, \bibinfo{person}{Shean Wang}, \bibinfo{person}{Lu Wang}, {and} \bibinfo{person}{Weizhu Chen}.} \bibinfo{year}{2021}\natexlab{}.
\newblock \showarticletitle{Lora: Low-rank adaptation of large language models}.
\newblock \bibinfo{journal}{\emph{arXiv preprint arXiv:2106.09685}} (\bibinfo{year}{2021}).
\newblock


\bibitem[Huang et~al\mbox{.}(2023)]%
        {huang2023ceval}
\bibfield{author}{\bibinfo{person}{Yuzhen Huang}, \bibinfo{person}{Yuzhuo Bai}, \bibinfo{person}{Zhihao Zhu}, \bibinfo{person}{Junlei Zhang}, \bibinfo{person}{Jinghan Zhang}, \bibinfo{person}{Tangjun Su}, \bibinfo{person}{Junteng Liu}, \bibinfo{person}{Chuancheng Lv}, \bibinfo{person}{Yikai Zhang}, \bibinfo{person}{Jiayi Lei}, \bibinfo{person}{Yao Fu}, \bibinfo{person}{Maosong Sun}, {and} \bibinfo{person}{Junxian He}.} \bibinfo{year}{2023}\natexlab{}.
\newblock \showarticletitle{C-Eval: A Multi-Level Multi-Discipline Chinese Evaluation Suite for Foundation Models}. In \bibinfo{booktitle}{\emph{Advances in Neural Information Processing Systems}}.
\newblock


\bibitem[Kaplan et~al\mbox{.}(2020)]%
        {kaplan2020scaling}
\bibfield{author}{\bibinfo{person}{Jared Kaplan}, \bibinfo{person}{Sam McCandlish}, \bibinfo{person}{Tom Henighan}, \bibinfo{person}{Tom~B Brown}, \bibinfo{person}{Benjamin Chess}, \bibinfo{person}{Rewon Child}, \bibinfo{person}{Scott Gray}, \bibinfo{person}{Alec Radford}, \bibinfo{person}{Jeffrey Wu}, {and} \bibinfo{person}{Dario Amodei}.} \bibinfo{year}{2020}\natexlab{}.
\newblock \showarticletitle{Scaling laws for neural language models}.
\newblock \bibinfo{journal}{\emph{arXiv preprint arXiv:2001.08361}} (\bibinfo{year}{2020}).
\newblock


\bibitem[Koncel-Kedziorski et~al\mbox{.}(2023)]%
        {koncel2023bizbench}
\bibfield{author}{\bibinfo{person}{Rik Koncel-Kedziorski}, \bibinfo{person}{Michael Krumdick}, \bibinfo{person}{Viet Lai}, \bibinfo{person}{Varshini Reddy}, \bibinfo{person}{Charles Lovering}, {and} \bibinfo{person}{Chris Tanner}.} \bibinfo{year}{2023}\natexlab{}.
\newblock \showarticletitle{Bizbench: A quantitative reasoning benchmark for business and finance}.
\newblock \bibinfo{journal}{\emph{arXiv preprint arXiv:2311.06602}} (\bibinfo{year}{2023}).
\newblock


\bibitem[Lewis et~al\mbox{.}(2020)]%
        {lewis2020retrieval}
\bibfield{author}{\bibinfo{person}{Patrick Lewis}, \bibinfo{person}{Ethan Perez}, \bibinfo{person}{Aleksandra Piktus}, \bibinfo{person}{Fabio Petroni}, \bibinfo{person}{Vladimir Karpukhin}, \bibinfo{person}{Naman Goyal}, \bibinfo{person}{Heinrich K{\"u}ttler}, \bibinfo{person}{Mike Lewis}, \bibinfo{person}{Wen-tau Yih}, \bibinfo{person}{Tim Rockt{\"a}schel}, {et~al\mbox{.}}} \bibinfo{year}{2020}\natexlab{}.
\newblock \showarticletitle{Retrieval-augmented generation for knowledge-intensive nlp tasks}.
\newblock \bibinfo{journal}{\emph{Advances in Neural Information Processing Systems}}  \bibinfo{volume}{33} (\bibinfo{year}{2020}), \bibinfo{pages}{9459--9474}.
\newblock


\bibitem[Li et~al\mbox{.}(2024)]%
        {li2024urbangpt}
\bibfield{author}{\bibinfo{person}{Zhonghang Li}, \bibinfo{person}{Lianghao Xia}, \bibinfo{person}{Jiabin Tang}, \bibinfo{person}{Yong Xu}, \bibinfo{person}{Lei Shi}, \bibinfo{person}{Long Xia}, \bibinfo{person}{Dawei Yin}, {and} \bibinfo{person}{Chao Huang}.} \bibinfo{year}{2024}\natexlab{}.
\newblock \showarticletitle{Urbangpt: Spatio-temporal large language models}.
\newblock \bibinfo{journal}{\emph{arXiv preprint arXiv:2403.00813}} (\bibinfo{year}{2024}).
\newblock


\bibitem[Liang et~al\mbox{.}(2022)]%
        {liang2022holistic}
\bibfield{author}{\bibinfo{person}{Percy Liang}, \bibinfo{person}{Rishi Bommasani}, \bibinfo{person}{Tony Lee}, \bibinfo{person}{Dimitris Tsipras}, \bibinfo{person}{Dilara Soylu}, \bibinfo{person}{Michihiro Yasunaga}, \bibinfo{person}{Yian Zhang}, \bibinfo{person}{Deepak Narayanan}, \bibinfo{person}{Yuhuai Wu}, \bibinfo{person}{Ananya Kumar}, {et~al\mbox{.}}} \bibinfo{year}{2022}\natexlab{}.
\newblock \showarticletitle{Holistic evaluation of language models}.
\newblock \bibinfo{journal}{\emph{arXiv preprint arXiv:2211.09110}} (\bibinfo{year}{2022}).
\newblock


\bibitem[Liu et~al\mbox{.}(2024)]%
        {liu2024mathbench}
\bibfield{author}{\bibinfo{person}{Hongwei Liu}, \bibinfo{person}{Zilong Zheng}, \bibinfo{person}{Yuxuan Qiao}, \bibinfo{person}{Haodong Duan}, \bibinfo{person}{Zhiwei Fei}, \bibinfo{person}{Fengzhe Zhou}, \bibinfo{person}{Wenwei Zhang}, \bibinfo{person}{Songyang Zhang}, \bibinfo{person}{Dahua Lin}, {and} \bibinfo{person}{Kai Chen}.} \bibinfo{year}{2024}\natexlab{}.
\newblock \showarticletitle{MathBench: Evaluating the Theory and Application Proficiency of LLMs with a Hierarchical Mathematics Benchmark}.
\newblock \bibinfo{journal}{\emph{arXiv preprint arXiv:2405.12209}} (\bibinfo{year}{2024}).
\newblock


\bibitem[Meta(2024)]%
        {llama}
\bibfield{author}{\bibinfo{person}{Meta}.} \bibinfo{year}{2024}\natexlab{}.
\newblock \bibinfo{title}{Introducing Meta Llama 3: The most capable openly available LLM to date}.
\newblock \bibinfo{howpublished}{\url{https://ai.meta.com/blog/meta-llama-3/}}.
\newblock


\bibitem[OpenAI(2022)]%
        {chatgpt}
\bibfield{author}{\bibinfo{person}{OpenAI}.} \bibinfo{year}{2022}\natexlab{}.
\newblock \bibinfo{title}{Chat{GPT}}.
\newblock \bibinfo{howpublished}{\url{https://chat.openai.com}}.
\newblock


\bibitem[Patterson(2012)]%
        {patterson2012better}
\bibfield{author}{\bibinfo{person}{David Patterson}.} \bibinfo{year}{2012}\natexlab{}.
\newblock \showarticletitle{For better or worse, benchmarks shape a field}.
\newblock \bibinfo{journal}{\emph{Commun. ACM}}  \bibinfo{volume}{55} (\bibinfo{year}{2012}).
\newblock


\bibitem[Qin et~al\mbox{.}(2024)]%
        {qin2024text2city}
\bibfield{author}{\bibinfo{person}{Yiming Qin}, \bibinfo{person}{Nanxuan Zhao}, \bibinfo{person}{Bin Sheng}, {and} \bibinfo{person}{Rynson~WH Lau}.} \bibinfo{year}{2024}\natexlab{}.
\newblock \showarticletitle{Text2City: One-Stage Text-Driven Urban Layout Regeneration}. In \bibinfo{booktitle}{\emph{Proceedings of the AAAI Conference on Artificial Intelligence}}, Vol.~\bibinfo{volume}{38}. \bibinfo{pages}{4578--4586}.
\newblock


\bibitem[Srivastava et~al\mbox{.}(2022)]%
        {srivastava2022beyond}
\bibfield{author}{\bibinfo{person}{Aarohi Srivastava}, \bibinfo{person}{Abhinav Rastogi}, \bibinfo{person}{Abhishek Rao}, \bibinfo{person}{Abu Awal~Md Shoeb}, \bibinfo{person}{Abubakar Abid}, \bibinfo{person}{Adam Fisch}, \bibinfo{person}{Adam~R Brown}, \bibinfo{person}{Adam Santoro}, \bibinfo{person}{Aditya Gupta}, \bibinfo{person}{Adri{\`a} Garriga-Alonso}, {et~al\mbox{.}}} \bibinfo{year}{2022}\natexlab{}.
\newblock \showarticletitle{Beyond the imitation game: Quantifying and extrapolating the capabilities of language models}.
\newblock \bibinfo{journal}{\emph{arXiv preprint arXiv:2206.04615}} (\bibinfo{year}{2022}).
\newblock


\bibitem[Su et~al\mbox{.}(2024a)]%
        {su2024large}
\bibfield{author}{\bibinfo{person}{Hongyuan Su}, \bibinfo{person}{Yu Zheng}, \bibinfo{person}{Jingtao Ding}, \bibinfo{person}{Depeng Jin}, {and} \bibinfo{person}{Yong Li}.} \bibinfo{year}{2024}\natexlab{a}.
\newblock \showarticletitle{Large-scale Urban Facility Location Selection with Knowledge-informed Reinforcement Learning}. In \bibinfo{booktitle}{\emph{Proceedings of the 32nd ACM International Conference on Advances in Geographic Information Systems}}. \bibinfo{pages}{553--556}.
\newblock


\bibitem[Su et~al\mbox{.}(2024b)]%
        {su2024metrognn}
\bibfield{author}{\bibinfo{person}{Hongyuan Su}, \bibinfo{person}{Yu Zheng}, \bibinfo{person}{Jingtao Ding}, \bibinfo{person}{Depeng Jin}, {and} \bibinfo{person}{Yong Li}.} \bibinfo{year}{2024}\natexlab{b}.
\newblock \showarticletitle{MetroGNN: Metro Network Expansion with Reinforcement Learning}. In \bibinfo{booktitle}{\emph{Companion Proceedings of the ACM on Web Conference 2024}}. \bibinfo{pages}{650--653}.
\newblock


\bibitem[Team et~al\mbox{.}(2024a)]%
        {team2024gemma}
\bibfield{author}{\bibinfo{person}{Gemma Team}, \bibinfo{person}{Thomas Mesnard}, \bibinfo{person}{Cassidy Hardin}, \bibinfo{person}{Robert Dadashi}, \bibinfo{person}{Surya Bhupatiraju}, \bibinfo{person}{Shreya Pathak}, \bibinfo{person}{Laurent Sifre}, \bibinfo{person}{Morgane Rivi{\`e}re}, \bibinfo{person}{Mihir~Sanjay Kale}, \bibinfo{person}{Juliette Love}, {et~al\mbox{.}}} \bibinfo{year}{2024}\natexlab{a}.
\newblock \showarticletitle{Gemma: Open models based on gemini research and technology}.
\newblock \bibinfo{journal}{\emph{arXiv preprint arXiv:2403.08295}} (\bibinfo{year}{2024}).
\newblock


\bibitem[Team et~al\mbox{.}(2024b)]%
        {team2024gemma2}
\bibfield{author}{\bibinfo{person}{Gemma Team}, \bibinfo{person}{Morgane Riviere}, \bibinfo{person}{Shreya Pathak}, \bibinfo{person}{Pier~Giuseppe Sessa}, \bibinfo{person}{Cassidy Hardin}, \bibinfo{person}{Surya Bhupatiraju}, \bibinfo{person}{L{\'e}onard Hussenot}, \bibinfo{person}{Thomas Mesnard}, \bibinfo{person}{Bobak Shahriari}, \bibinfo{person}{Alexandre Ram{\'e}}, {et~al\mbox{.}}} \bibinfo{year}{2024}\natexlab{b}.
\newblock \showarticletitle{Gemma 2: Improving open language models at a practical size}.
\newblock \bibinfo{journal}{\emph{arXiv preprint arXiv:2408.00118}} (\bibinfo{year}{2024}).
\newblock


\bibitem[Touvron et~al\mbox{.}(2023)]%
        {touvron2023llama}
\bibfield{author}{\bibinfo{person}{Hugo Touvron}, \bibinfo{person}{Thibaut Lavril}, \bibinfo{person}{Gautier Izacard}, \bibinfo{person}{Xavier Martinet}, \bibinfo{person}{Marie-Anne Lachaux}, \bibinfo{person}{Timoth{\'e}e Lacroix}, \bibinfo{person}{Baptiste Rozi{\`e}re}, \bibinfo{person}{Naman Goyal}, \bibinfo{person}{Eric Hambro}, \bibinfo{person}{Faisal Azhar}, {et~al\mbox{.}}} \bibinfo{year}{2023}\natexlab{}.
\newblock \showarticletitle{Llama: Open and efficient foundation language models}.
\newblock \bibinfo{journal}{\emph{arXiv preprint arXiv:2302.13971}} (\bibinfo{year}{2023}).
\newblock


\bibitem[Wang et~al\mbox{.}(2019)]%
        {wang2019superglue}
\bibfield{author}{\bibinfo{person}{Alex Wang}, \bibinfo{person}{Yada Pruksachatkun}, \bibinfo{person}{Nikita Nangia}, \bibinfo{person}{Amanpreet Singh}, \bibinfo{person}{Julian Michael}, \bibinfo{person}{Felix Hill}, \bibinfo{person}{Omer Levy}, {and} \bibinfo{person}{Samuel Bowman}.} \bibinfo{year}{2019}\natexlab{}.
\newblock \showarticletitle{Superglue: A stickier benchmark for general-purpose language understanding systems}.
\newblock \bibinfo{journal}{\emph{Advances in neural information processing systems}}  \bibinfo{volume}{32} (\bibinfo{year}{2019}).
\newblock


\bibitem[Wang et~al\mbox{.}(2020)]%
        {wang2020reimagining}
\bibfield{author}{\bibinfo{person}{Dongjie Wang}, \bibinfo{person}{Yanjie Fu}, \bibinfo{person}{Pengyang Wang}, \bibinfo{person}{Bo Huang}, {and} \bibinfo{person}{Chang-Tien Lu}.} \bibinfo{year}{2020}\natexlab{}.
\newblock \showarticletitle{Reimagining city configuration: Automated urban planning via adversarial learning}. In \bibinfo{booktitle}{\emph{Proceedings of the 28th international conference on advances in geographic information systems}}. \bibinfo{pages}{497--506}.
\newblock


\bibitem[Wang et~al\mbox{.}(2024)]%
        {wang2024transgpt}
\bibfield{author}{\bibinfo{person}{Peng Wang}, \bibinfo{person}{Xiang Wei}, \bibinfo{person}{Fangxu Hu}, {and} \bibinfo{person}{Wenjuan Han}.} \bibinfo{year}{2024}\natexlab{}.
\newblock \showarticletitle{TransGPT: Multi-modal Generative Pre-trained Transformer for Transportation}.
\newblock \bibinfo{journal}{\emph{arXiv preprint arXiv:2402.07233}} (\bibinfo{year}{2024}).
\newblock


\bibitem[Wei et~al\mbox{.}(2022a)]%
        {wei2022emergent}
\bibfield{author}{\bibinfo{person}{Jason Wei}, \bibinfo{person}{Yi Tay}, \bibinfo{person}{Rishi Bommasani}, \bibinfo{person}{Colin Raffel}, \bibinfo{person}{Barret Zoph}, \bibinfo{person}{Sebastian Borgeaud}, \bibinfo{person}{Dani Yogatama}, \bibinfo{person}{Maarten Bosma}, \bibinfo{person}{Denny Zhou}, \bibinfo{person}{Donald Metzler}, {et~al\mbox{.}}} \bibinfo{year}{2022}\natexlab{a}.
\newblock \showarticletitle{Emergent abilities of large language models}.
\newblock \bibinfo{journal}{\emph{arXiv preprint arXiv:2206.07682}} (\bibinfo{year}{2022}).
\newblock


\bibitem[Wei et~al\mbox{.}(2022b)]%
        {wei2022chain}
\bibfield{author}{\bibinfo{person}{Jason Wei}, \bibinfo{person}{Xuezhi Wang}, \bibinfo{person}{Dale Schuurmans}, \bibinfo{person}{Maarten Bosma}, \bibinfo{person}{Fei Xia}, \bibinfo{person}{Ed Chi}, \bibinfo{person}{Quoc~V Le}, \bibinfo{person}{Denny Zhou}, {et~al\mbox{.}}} \bibinfo{year}{2022}\natexlab{b}.
\newblock \showarticletitle{Chain-of-thought prompting elicits reasoning in large language models}.
\newblock \bibinfo{journal}{\emph{Advances in neural information processing systems}}  \bibinfo{volume}{35} (\bibinfo{year}{2022}), \bibinfo{pages}{24824--24837}.
\newblock


\bibitem[Xu et~al\mbox{.}(2023)]%
        {xu2023urban}
\bibfield{author}{\bibinfo{person}{Fengli Xu}, \bibinfo{person}{Jun Zhang}, \bibinfo{person}{Chen Gao}, \bibinfo{person}{Jie Feng}, {and} \bibinfo{person}{Yong Li}.} \bibinfo{year}{2023}\natexlab{}.
\newblock \showarticletitle{Urban generative intelligence (ugi): A foundational platform for agents in embodied city environment}.
\newblock \bibinfo{journal}{\emph{arXiv preprint arXiv:2312.11813}} (\bibinfo{year}{2023}).
\newblock


\bibitem[Yang et~al\mbox{.}(2024)]%
        {yang2024qwen2}
\bibfield{author}{\bibinfo{person}{An Yang}, \bibinfo{person}{Baosong Yang}, \bibinfo{person}{Binyuan Hui}, \bibinfo{person}{Bo Zheng}, \bibinfo{person}{Bowen Yu}, \bibinfo{person}{Chang Zhou}, \bibinfo{person}{Chengpeng Li}, \bibinfo{person}{Chengyuan Li}, \bibinfo{person}{Dayiheng Liu}, \bibinfo{person}{Fei Huang}, {et~al\mbox{.}}} \bibinfo{year}{2024}\natexlab{}.
\newblock \showarticletitle{Qwen2 technical report}.
\newblock \bibinfo{journal}{\emph{arXiv preprint arXiv:2407.10671}} (\bibinfo{year}{2024}).
\newblock


\bibitem[Young et~al\mbox{.}(2024)]%
        {young2024yi}
\bibfield{author}{\bibinfo{person}{Alex Young}, \bibinfo{person}{Bei Chen}, \bibinfo{person}{Chao Li}, \bibinfo{person}{Chengen Huang}, \bibinfo{person}{Ge Zhang}, \bibinfo{person}{Guanwei Zhang}, \bibinfo{person}{Heng Li}, \bibinfo{person}{Jiangcheng Zhu}, \bibinfo{person}{Jianqun Chen}, \bibinfo{person}{Jing Chang}, {et~al\mbox{.}}} \bibinfo{year}{2024}\natexlab{}.
\newblock \showarticletitle{Yi: Open foundation models by 01. ai}.
\newblock \bibinfo{journal}{\emph{arXiv preprint arXiv:2403.04652}} (\bibinfo{year}{2024}).
\newblock


\bibitem[Zeng et~al\mbox{.}(2023)]%
        {zeng2023glm-130b}
\bibfield{author}{\bibinfo{person}{Aohan Zeng}, \bibinfo{person}{Xiao Liu}, \bibinfo{person}{Zhengxiao Du}, \bibinfo{person}{Zihan Wang}, \bibinfo{person}{Hanyu Lai}, \bibinfo{person}{Ming Ding}, \bibinfo{person}{Zhuoyi Yang}, \bibinfo{person}{Yifan Xu}, \bibinfo{person}{Wendi Zheng}, \bibinfo{person}{Xiao Xia}, \bibinfo{person}{Weng~Lam Tam}, \bibinfo{person}{Zixuan Ma}, \bibinfo{person}{Yufei Xue}, \bibinfo{person}{Jidong Zhai}, \bibinfo{person}{Wenguang Chen}, \bibinfo{person}{Zhiyuan Liu}, \bibinfo{person}{Peng Zhang}, \bibinfo{person}{Yuxiao Dong}, {and} \bibinfo{person}{Jie Tang}.} \bibinfo{year}{2023}\natexlab{}.
\newblock \showarticletitle{{GLM}-130B: An Open Bilingual Pre-trained Model}. In \bibinfo{booktitle}{\emph{The Eleventh International Conference on Learning Representations (ICLR)}}.
\newblock
\urldef\tempurl%
\url{https://openreview.net/forum?id=-Aw0rrrPUF}
\showURL{%
\tempurl}


\bibitem[Zhang et~al\mbox{.}(2024)]%
        {zhang2024trafficgpt}
\bibfield{author}{\bibinfo{person}{Siyao Zhang}, \bibinfo{person}{Daocheng Fu}, \bibinfo{person}{Wenzhe Liang}, \bibinfo{person}{Zhao Zhang}, \bibinfo{person}{Bin Yu}, \bibinfo{person}{Pinlong Cai}, {and} \bibinfo{person}{Baozhen Yao}.} \bibinfo{year}{2024}\natexlab{}.
\newblock \showarticletitle{Trafficgpt: Viewing, processing and interacting with traffic foundation models}.
\newblock \bibinfo{journal}{\emph{Transport Policy}}  \bibinfo{volume}{150} (\bibinfo{year}{2024}), \bibinfo{pages}{95--105}.
\newblock


\bibitem[Zhang et~al\mbox{.}(2023)]%
        {zhang2023siren}
\bibfield{author}{\bibinfo{person}{Yue Zhang}, \bibinfo{person}{Yafu Li}, \bibinfo{person}{Leyang Cui}, \bibinfo{person}{Deng Cai}, \bibinfo{person}{Lemao Liu}, \bibinfo{person}{Tingchen Fu}, \bibinfo{person}{Xinting Huang}, \bibinfo{person}{Enbo Zhao}, \bibinfo{person}{Yu Zhang}, \bibinfo{person}{Yulong Chen}, {et~al\mbox{.}}} \bibinfo{year}{2023}\natexlab{}.
\newblock \showarticletitle{Siren's song in the AI ocean: a survey on hallucination in large language models}.
\newblock \bibinfo{journal}{\emph{arXiv preprint arXiv:2309.01219}} (\bibinfo{year}{2023}).
\newblock


\bibitem[Zheng et~al\mbox{.}(2024a)]%
        {zheng2024survey}
\bibfield{author}{\bibinfo{person}{Yu Zheng}, \bibinfo{person}{Qianyue Hao}, \bibinfo{person}{Jingwei Wang}, \bibinfo{person}{Changzheng Gao}, \bibinfo{person}{Jinwei Chen}, \bibinfo{person}{Depeng Jin}, {and} \bibinfo{person}{Yong Li}.} \bibinfo{year}{2024}\natexlab{a}.
\newblock \showarticletitle{A Survey of Machine Learning for Urban Decision Making: Applications in Planning, Transportation, and Healthcare}.
\newblock \bibinfo{journal}{\emph{Comput. Surveys}} \bibinfo{volume}{57}, \bibinfo{number}{4} (\bibinfo{year}{2024}), \bibinfo{pages}{1--41}.
\newblock


\bibitem[Zheng et~al\mbox{.}(2023a)]%
        {zheng2023spatial}
\bibfield{author}{\bibinfo{person}{Yu Zheng}, \bibinfo{person}{Yuming Lin}, \bibinfo{person}{Liang Zhao}, \bibinfo{person}{Tinghai Wu}, \bibinfo{person}{Depeng Jin}, {and} \bibinfo{person}{Yong Li}.} \bibinfo{year}{2023}\natexlab{a}.
\newblock \showarticletitle{Spatial planning of urban communities via deep reinforcement learning}.
\newblock \bibinfo{journal}{\emph{Nature Computational Science}} \bibinfo{volume}{3}, \bibinfo{number}{9} (\bibinfo{year}{2023}), \bibinfo{pages}{748--762}.
\newblock


\bibitem[Zheng et~al\mbox{.}(2023b)]%
        {zheng2023road}
\bibfield{author}{\bibinfo{person}{Yu Zheng}, \bibinfo{person}{Hongyuan Su}, \bibinfo{person}{Jingtao Ding}, \bibinfo{person}{Depeng Jin}, {and} \bibinfo{person}{Yong Li}.} \bibinfo{year}{2023}\natexlab{b}.
\newblock \showarticletitle{Road planning for slums via deep reinforcement learning}. In \bibinfo{booktitle}{\emph{Proceedings of the 29th ACM SIGKDD Conference on Knowledge Discovery and Data Mining}}. \bibinfo{pages}{5695--5706}.
\newblock


\bibitem[Zheng et~al\mbox{.}(2024b)]%
        {zheng2024llamafactory}
\bibfield{author}{\bibinfo{person}{Yaowei Zheng}, \bibinfo{person}{Richong Zhang}, \bibinfo{person}{Junhao Zhang}, \bibinfo{person}{Yanhan Ye}, \bibinfo{person}{Zheyan Luo}, {and} \bibinfo{person}{Yongqiang Ma}.} \bibinfo{year}{2024}\natexlab{b}.
\newblock \showarticletitle{LlamaFactory: Unified Efficient Fine-Tuning of 100+ Language Models}.
\newblock \bibinfo{journal}{\emph{arXiv preprint arXiv:2403.13372}} (\bibinfo{year}{2024}).
\newblock
\urldef\tempurl%
\url{http://arxiv.org/abs/2403.13372}
\showURL{%
\tempurl}


\bibitem[Zhou et~al\mbox{.}(2024)]%
        {zhou2024large}
\bibfield{author}{\bibinfo{person}{Zhilun Zhou}, \bibinfo{person}{Yuming Lin}, \bibinfo{person}{Depeng Jin}, {and} \bibinfo{person}{Yong Li}.} \bibinfo{year}{2024}\natexlab{}.
\newblock \showarticletitle{Large Language Model for Participatory Urban Planning}.
\newblock \bibinfo{journal}{\emph{arXiv preprint arXiv:2402.17161}} (\bibinfo{year}{2024}).
\newblock


\bibitem[Zhu et~al\mbox{.}(2024)]%
        {zhu2024plangpt}
\bibfield{author}{\bibinfo{person}{He Zhu}, \bibinfo{person}{Wenjia Zhang}, \bibinfo{person}{Nuoxian Huang}, \bibinfo{person}{Boyang Li}, \bibinfo{person}{Luyao Niu}, \bibinfo{person}{Zipei Fan}, \bibinfo{person}{Tianle Lun}, \bibinfo{person}{Yicheng Tao}, \bibinfo{person}{Junyou Su}, \bibinfo{person}{Zhaoya Gong}, {et~al\mbox{.}}} \bibinfo{year}{2024}\natexlab{}.
\newblock \showarticletitle{PlanGPT: Enhancing Urban Planning with Tailored Language Model and Efficient Retrieval}.
\newblock \bibinfo{journal}{\emph{arXiv preprint arXiv:2402.19273}} (\bibinfo{year}{2024}).
\newblock


\end{thebibliography}

\appendix

\section{Case Study}\label{app:case}

In real-world urban planning scenarios, human planners spend much time dealing with textual data. We show two typical cases about urban planning text polishment and proofreading, to demonstrate the effectiveness of our tool in practical applications.

\noindent\textbf{Urban planning text polishment}

\noindent\verb|#prompt#|
Assuming that you are a staff member who is good at writing and proofreading China's urban and rural planning texts, could you please polish and proofread the \#original text\# according to the \#urban and rural planning text requirements\#: 

\noindent\verb|#urban and rural planning text requirements#|
The text is coherent and complete.
Standardized terminology, official and rigorous content, no colloquial expressions, no English, pinyin, internet terms, etc.
In line with the style of government documents, the content is full, and there is no concept stacking and repetition of content.
The content is reasonable and accurate, no misleading expressions, ensure consistency and logic.
Coorect in grammar and spelling.

\noindent\verb|#original text#|
It is difficult to find a historical and cultural district worthy of protection in a famous historical and cultural city, and the focus now is on restoring the traditional appearance of the historical city center as a whole.

\noindent\verb|#response#|
A famous historical and cultural city currently has difficulty in finding a historical and cultural neighborhood worthy of protection. In order to enhance the historical and cultural value of the city, measures should be taken to enhance the protection of the existing cultural heritage, to identify and discover historical and cultural districts that have yet to be protected, and to restore the traditional appearance of the historical city center as a whole, so as to preserve and continue the historical and cultural characteristics of the city.

\noindent\textbf{Urban planning text proofreading}

\noindent\verb|#prompt#|
Assuming that you are a staff member who is good at writing and proofreading Chinese town and country planning texts, please touch up and proofread the \#original text\# according to the \#urban and rural planning text requirements\#. 

\noindent\verb|#urban and rural planning text requirements#|
The text is coherent and complete.
Standardised terminology, official and rigorous content, no colloquial expressions, no English, pinyin, internet terms, etc.
In line with the style of government documents, beautiful language, full of content, without concept stacking and repetitive content.
Whether the content is reasonable and accurate, whether there are misleading expressions, to ensure consistency and logic.
Pay attention to grammar, spelling and fluency.

\noindent\verb|#original text#|
The layout of the city of Athens is a complete embodiment of the Hippodrome layout pattern, the Miletus is a layout pattern centred on the city square, with a grid of roads as the backbone , the square or the marketplace is surrounded by a series of public buildings, which are the core of urban life, the Acropolis has a very typical unplanned layout of features

\noindent\verb|#response#|
The layout of the Athenian city does not fully reflect the Hippodrome layout pattern, while the Miletus is a typical layout pattern centred on the city square, with a grid of roads as the skeleton. A series of public buildings were constructed around the square or marketplace, which became the core of urban life. In addition, the Acropolis exhibits a typical unplanned layout characteristic, reflecting the diversity of history and culture.

\verb|(The above contents are translated from Chinese)|

\section{Prompt for data generation}\label{app:prompt}
The prompt template of data generation is as follows:

\verb|#prompt|

\#01 You are a Q\&A pair dataset processing expert.

\#02 Your task is to generate corresponding Q\&A pairs based on my questions and the content I give.

\#03 The questions generated must be macro and value based, don't generate particularly detailed questions, and not too long.

\#04 Answers must be comprehensive, use more of my information and be more informative.

\#05 The text is coherent and the content is complete.

\#06 Use standardised language, official and rigorous content, no colloquial expressions, no English, pinyin, internet terms, etc.

\#07 Beautiful language, rigorous content, no concepts and repetitive content.

\#08 Must be generated according to the following sample format:

{

"instruction": “question”,

"output": "answer"

}

\#09 The reference case is as follows:

{

"instruction": "How can town system planning contribute to the sustainable development of a town system?" ,

"output": "Town system planning to promote the sustainable development of the town system needs to consider comprehensively from the aspects of resource utilisation, environmental protection, social and economic development."

}

Generate 20 Q\&A pairs, ending with the full JSON formatted Q\&A pair, discarding contexts that do not respond to completion.

\verb|(The above contents are translated from Chinese)|

\end{document}